\documentclass[preprint,12pt,authoryear]{elsarticle}
\usepackage{amssymb}
\usepackage{amsmath}
\usepackage{natbib}  % 关键：必须加载
\usepackage{multirow}%
\usepackage{booktabs}%
\usepackage{amsfonts}%
\usepackage{caption}
\usepackage{url}
\usepackage{lineno}
\usepackage{xcolor}  % 引入颜色支持
\journal{Artificial Intelligence in Agriculture}
\begin{document}
% \linenumbers
\renewcommand{\textcolor}[2]{#2}

\begin{frontmatter}

\cortext[cor1]{Corresponding authors}

\title{Vision-Based Early Fault Diagnosis and Self-Recovery for Strawberry Harvesting Robots} %% Article title

\author[1]{Meili Sun} %\email{sunmeili@stu.shzu.edu.cn} %% Author name 
\author[1,2]{Chunjiang Zhao \corref{cor1}} %% Author name
\ead{zhaocj@nercita.org.cn}

\author[2]{Lichao Yang} %% Author name
\author[2]{Hao Liu} %% Author name
\author[2]{Shimin Hu} %% Author name
\author[1,2]{Ya Xiong \corref{cor1}} %% Author name 
\ead{yaxiong@nercita.org.cn}

\affiliation[1]{organization={College of Mechanical and Electrical Engineering, Shihezi University},%Department and Organization
            % addressline={}, 
            city={Xinjiang},
            postcode={832003}, 
            % state={},
            country={China}}
%% Author affiliation
\affiliation[2]{organization={Intelligent Equipment Research Center, Beijing Academy of Agriculture and Forestry Sciences},%Department and Organization
            % addressline={}, 
            city={Beijing},
            postcode={100097}, 
            % state={},
            country={China}}

%% Abstract
\begin{abstract}

\textcolor{red}{Strawberry-harvesting robots faced challenges such as poor visual perception, gripper misalignment, empty grasp/misgrasp, and slippage, which reduced harvesting stability and efficiency.
To overcome these issues, this paper proposes a visual fault diagnosis and self-recovery framework. An end-to-end SRR-Net achieved unified perception and fault diagnosis through joint detection, segmentation, and ripeness regression of the fruit and gripper. Leveraging this integrated perception, a relative error compensation method driven by simultaneous target-gripper detection was designed to correct positional misalignments exceeding the tolerance threshold. A micro-optical camera integrated within the end-effector delivered real-time visual feedback. Based on the micro-optical camera, a MobileNet V3-Small classifier was utilized for grasp adjustment during the deflating stage, enabling the early abort of the harvesting cycle in cases of empty grasp/misgrasps. Furthermore, a time-series LSTM classifier was applied during the snap-off stage to predict strawberry slippage. Based on these predictions, the system executed re-inflation and a secondary snap-off attempt for slipping strawberries, or aborted the cycle for slipped strawberries. Experiments demonstrated that the mean absolute errors between the end-effector and the picking point were reduced to 3.12 mm and \textcolor{blue}{4.06 mm from 11.50 mm and 5.25 mm along the $x$- and $y$-axes, respectively, at the cost of a time increment of 0.64 $\pm$ 0.24 s}. The grasp adjustment module reduced the grasping phase by approximately 0.5 s and avoided empty-placement for failure cases. The strawberry slip prediction module handled slipped cases with an 88.89\% success rate, saving approximately 4.00 s per harvesting cycle for failure cases. Also, it achieved an 81.25\% recovery rate for slipping strawberries, requiring additional 0.63 s for re-grasping. Overall, the framework may establish a highly practical solution for autonomous fruit harvesting, supported by a video demonstration of visual diagnosis and self-recovery at \url{https://youtu.be/UOfwlHgXUgU}.}

\end{abstract}

%%Research highlights
\begin{highlights}
\item Integrated vision perception and fault feedback into a unified end-to-end multi-task perception framework.
\item Proposed a visual \textcolor{red}{relative} error compensation method to accurately align picking point with gripper.
\item Designed an early abort fault diagnosis and prediction method to improve efficiency.
\end{highlights}


\end{frontmatter}

\section{Introduction}

Harvesting robots have made significant progress in recent years, showing great potential to reduce labor dependence and improve agricultural productivity \citep{hua2025key,xiao2024review}. \textcolor{blue}{However, several mechanical, electrical, and control faults \citep{milecki2023review, rajendran2024towards}} \textcolor{red}{have arisen} during robotic operations, compromising operational stability and continuity. For example, mechanical structural failures \citep{wang2025structural}, air leakage in pneumatic end-effectors \citep{rapalo2024design}, and fractures in end-joint connectors \textcolor{red}{have impaired} the normal operation of strawberry harvesting robots. Moreover, the absence of active learning and self-updating mechanisms \textcolor{red}{has rendered} models ineffective in adapting to changing conditions. Failure to respond to abnormal signals \textcolor{red}{has also posed} operational and safety risks to the robot. Once any single component \textcolor{red}{failed or produced} an erroneous output, the entire harvesting process \textcolor{red}{was} interrupted. Due to the lack of robust fault diagnosis and self-recovery mechanisms, harvesting robots \textcolor{red}{remained} prone to frequent work interruptions, which \textcolor{red}{shortened} their effective operation time in the field. 

Generally, fruit harvesting with robots \textcolor{red}{involved} seven key steps: image acquisition, fruit detection, segmentation and ripeness estimation, instance tracking and localization, motion planning, execution, result evaluation with fault diagnosis, and result recording with optimization. Along this pipeline, potential faults \textcolor{red}{occurred} in multiple stages \citep{khalastchi2019fault}. \textcolor{blue}{On the perception side, problems such as blurred or occluded cameras, unstable illumination \citep{hou2023overview}}, network/data loss, inaccurate segmentation \citep{shi2025advances, ZHANG2025539}, ripeness misclassification \citep{kamat2025multi}, ID tracking drift, and depth or calibration errors can lead to detection failures. At the motion control level, incorrect path planning, collision avoidance errors, inverse kinematics failures, or grasp misalignment \textcolor{red}{disrupted} harvesting. Even after successful contact, execution faults such as loose grasps, failed detachments, fruit damage, or gripper jamming \textcolor{red}{remained} common. These \textcolor{red}{were} further compounded by inaccurate success/failure judgment and inadequate fault recovery, resulting in prolonged downtime. Without robust fault diagnosis and self-recovery mechanisms, current robots \textcolor{red}{were} prone to frequent interruptions, severely limiting their effective operating time in orchards. 

% Figure
\begin{figure}%[]
  \centering
    \includegraphics[width=1.12in]{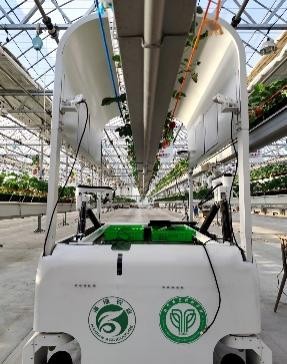}
    \includegraphics[width=1.83in]{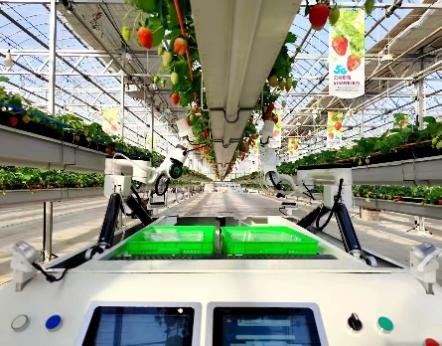}
    \includegraphics[width=1.88in]{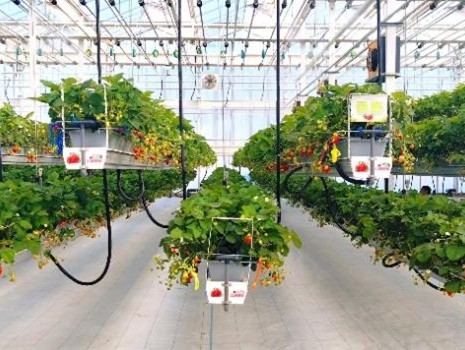}
    \caption{Images collection devices and environment. (a) HarvestFlex robot with a shading cover; (b) HarvestFlex robot without a shading cover; (c) table-top strawberries}\label{fig-1}
\end{figure}

% Figure
\begin{figure*}%[]
  \centering
   \includegraphics[width=0.8in]{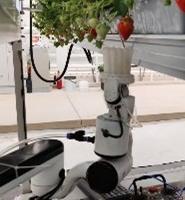}
   \includegraphics[width=0.8in]{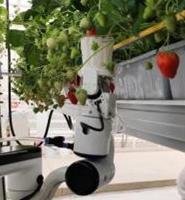}
   \includegraphics[width=0.8in]{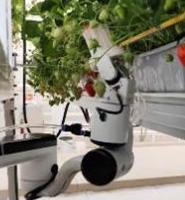}
   \includegraphics[width=0.8in]{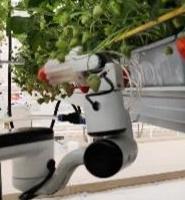}
   \includegraphics[width=0.8in]{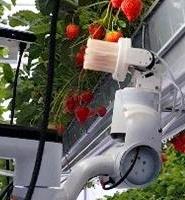}
   \includegraphics[width=0.8in]{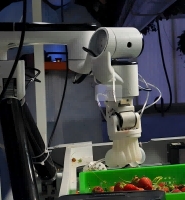}
    \caption{The picking process of strawberry harvesting robot: inflating and approaching, swallowing, deflating, snap off, descending and placing.}\label{fig-3}
\end{figure*}
In the HarvestFlex system (https://xiong-lab.cn/), as shown in Fig. \ref{fig-1} (a) and (b), a RealSense D455F depth camera (Intel, Realsense, USA) captured real-time RGB-D images, which were processed by YOLOv11 for strawberry detection and segmentation. A checkerboard-based calibration enabled the initial eye-to-hand coordinate transformation, and a regression method further compensated for spatial errors \textcolor{red}{\citep{hu2026calibration}}. For sequence planning, a minimal-height sorting algorithm determined harvesting order, and a point-to-point speed pattern-based method ensured efficient \textcolor{red}{movement} the end-effector. The end-effector was a flexible pneumatic gripper \citep{11231358} that inflated and deflated to gently envelop and detach strawberries. The complete harvesting workflow consisted of six steps: inflating and approaching, swallowing, deflating, snap off, descending and placing, in Fig. \ref{fig-3}. 

However, several issues emerged with our HarvestFlex robot that severely compromised its ability to harvest stably and continuously. These primary challenges were: 
(1) low integration of visual perception tasks -- the limited fusion of detection, segmentation, and ripeness estimation led to increased model complexity and reduced inference speed, thereby constraining overall system efficiency. 
(2) Positional inaccuracy between the end-effector and the target fruit -- \textcolor{red}{this deviation was primarily caused by mechanical looseness, variations in ambient illumination, frequent disassembly and installation, and unexpected fruit motion.} Even minor deviations could lead to misalignment, empty grasps, unsuccessful detachments, or incorrect placements, thereby disrupting continuous operation. 
(3) Inadequate grasping and strawberry slippage -- although the pneumatic soft gripper reduced fruit damage, its high compliance \textcolor{red}{made} the integration of reliable tactile or slippage sensors challenging. Consequently, slippage often went undetected, resulting in wasted cycle time and, in some cases, unintended fruit release during the snap-off stage. This problem was further compounded by the immaturity of current tactile sensing technologies, which remained expensive, unstable, and impractical for large-scale deployment in orchard environments. Moreover, existing tactile sensor technologies remained immature, with solutions that \textcolor{red}{were} expensive, unstable, and \textcolor{red}{difficult} to deploy reliably in orchard environments.

To mitigate such problems, fault diagnosis and recovery methods were  proposed to deal with electrical, software and control, process-related, and human error faults \citep{sabry2024review}. Traditional methods include signal processing techniques \citep{zhang2022vibration}, rule-based systems \citep{vukadinovic2024anomaly}, model-based approaches \citep{hasan2023model}, etc. However, these methods faced limitations in handling complex fault scenarios, particularly in the case of rule-based and threshold-based approaches. With advances in computer hardware, deep learning emerged a new paradigm for fault diagnosis, enabling automatic feature extraction from multi-modal data such as sensor signals, vibrations, acoustics, and images. CNNs were applied to surface defect detection \citep{zhou2025finger}, while RNNs and LSTMs were used to analyze time-series data for early fault detection in \citep{hwang2023cooperation}. 
\textcolor{red}{
Despite these algorithmic advancements, the performance of such deep learning models heavily depended on the continuous acquisition of high-quality, multi-modal data. In robotic manipulation tasks, capturing precise physical states---particularly for identifying and compensating misalignments between the end-effector and the target fruit---often demanded the installation of dedicated monitoring devices.
While laser trackers or vision sensors can compensate relative error for the misalignment between the target and the gripper \citep{LI2023112418, ferrarini2024method}, these approaches inevitably necessitate the integration of additional sensor hardware. Furthermore, the inclusion of these external devices inherently introduced additional calibration errors into the system. Consequently, relying on such supplementary sensors to acquire high-quality multi-modal data not only increased overall cost and hardware complexity but also severely complicated their practical deployment in real-world agricultural environments.}
% However, these methods often relied on additional sensors to acquire high-quality multi-modal data, which not only increased system cost and complexity but also complicated deployment in real-world agricultural environments. 
Furthermore, ensuring temporal and spatial alignment across heterogeneous data sources (e.g., vision, vibration, and tactile signals) remained a challenge, as even slight misalignments could degrade fault diagnosis accuracy. 

To overcome these limitations, this paper focuses on a vision-based fault diagnosis framework for HarvestFlex, thereby avoiding the need for additional multi-modal sensors and the complexities of data alignment. The objective was to address software- and control-related faults that compromised harvesting stability and efficiency. To this end, an end-to-end multi-task perception method SRR-Net \citep{SUN2026111169} was introduced and integrated visual perception with fault object detection while maintaining a lightweight model architecture and low computational load. The method comprised three primary subtasks: object detection, instance segmentation, and ripeness \textcolor{red}{regression}. \textcolor{red}{To address gripper misalignment with respect to the harvesting point}, a relative error compensation method based on the simultaneous target-gripper detection was implemented in the coordinate frame of robot arm once the gripper reached the position beneath the target. \textcolor{red}{If the error exceeded the maximum tolerance threshold,} a corrected harvesting point was generated, guiding the gripper to move beneath the target for a second time to ensure accurate alignment.
Furthermore, an early abort strategy was introduced to improve the reliability of the harvesting process. Specifically, a micro-optical camera embedded at the base of the gripper continuously monitors the presence and stability of strawberries. In the deflating stage, the MobileNet V3-Small \citep{9008835} verified whether the fruit had been successfully grasped, 
while in the snap-off stage, an LSTM classifier estimated the probability of the strawberry slippage from the gripper. If the strawberry was slipping, re-inflation and snap-off actions were triggered. If it had already slipped, an early abort signal was issued to \textcolor{red}{abort} the current cycle. This integrated perception–action framework enabled prompt diagnosis of empty grasps/misgrasp, and slippage, allowing the robotic arm to respond rapidly and maintain stable operation. The main contributions of this paper were as follows. 
\begin{itemize}
\item An end-to-end multi-task perception framework, SRR-Net, was introduced to integrate visual perception with fault diagnosis. 
\item A relative error compensation method based on the simultaneous target-gripper detection was developed to realign the position of end-effector when approaching the area beneath the picking point. 
\item To enhance operational reliability, an early abort strategy was implemented. During the deflating stage, MobileNet V3-Small was used to classify whether the strawberry had been successfully grasped. Subsequently, in the snap-off stage, an LSTM classifier predicted the likelihood of fruit slippage from the gripper, enabling timely corrective actions. 
\end{itemize}

% Main text
\section{Dataset Benchmark}
In this paper, FaultData, GraspData, SnapData and SlipData were constructed to support the multi-task vision perception task, grasp adjustment task and enable the time-series LSTM classifier task. All data were collected using the HarvestFlex strawberry-harvesting robot, which includes two robotic arms, two Realsense D455F cameras, as shown in Fig. \ref{fig-1}(a–b). The cameras captured RGB-D images with the distance between the lens and the strawberries ranging from 30 to 90 cm. Data were collected in both natural and LED lighting environments to enhance the robot’s visual adaptability to complex and dynamic orchard conditions. 

\textbf{FaultData} For the tasks of detection, segmentation, and ripeness \textcolor{red}{regression} of strawberry and gripper, diverse strawberry formations-such as isolated, overlapping, and occluded fruits-as well as the complete operational sequence of the end-effector during picking images were collected. To ensure robust and continuous operation for multi-task vision perception task, the dataset also incorporated a variety of lighting conditions, weather scenarios, and nighttime environments, laying the foundation for stable, autonomous, all-weather performance. LightStrawberry from \citep{SUN2026111169} also was applied into FaultData. In addition, high-resolution images were captured using a Redmi Note 13 Pro smartphone and an Orbbec Gemini Pro RGB-D camera (Orbbec, Gemini Pro, China) providing enhanced texture and color information. The entire dataset was collected at the Cuihu Farms in Beijing, China, and featured the 
\textit{Fragaria × ananassa} `Kaorino' cultivar, as shown in Fig. \ref{fig-1}(c). All images were annotated with the polygonal outlines of strawberries and the end-effector using Labelme \citep{torralba2010labelme}. \textcolor{red}{The dataset comprised two distinct classes: strawberry and hand.} Strawberry with the ripeness of strawberries assigned based on \citep{SUN2026111169}  and hand were labeled. The ripeness score was defined within the range [0, 1.1]. The label format for each instance followed the structure: $<$\textit{cls}$>$ $<$\textit{ripeness}$>$ $<$\textit{boundaries}$>$, where \textit{cls} represented the object class. The dataset was split into training and validation subsets, consisting of 2954 and 779 images, respectively. Note that the ripeness attribute applied only to strawberries. To ensure a consistent label format, the hand class was assigned alignment flags of 2. 

\textbf{GraspData} \textcolor{red}{For the grasp adjustmetn,image data were collected during the deflation stage of the end-effector using a miniature camera.} The dataset comprised three classes: 389 images of ripe strawberrie held in the gripper (Class 0), 346 images of empty grasps with no strawberry (Class 1), and 388 images of unripe strawberries held in the gripper (Class 2). Following a 7:3 stratified split, the training set contained 272, 242, and 272 images for the three classes respectively, while the validation set contained 117, 104, and 116 images. \textcolor{red}{A prominent challenge was distinguishing the critical state where the strawberry was only partially enclosed by the gripper. The categorization of this state directly impacted the stability of the grasp adjustment. During the deflating phase, a transition of the pneumatic end-effector from an inflated state to a deflated state was observed. From the perspective of the camera at the base of the gripper, the scene shifted from including the external environment to only revealing the strawberry and internal gripper components. During this transition, a visual boundary state was formed, wherein the strawberry was only partially positioned inside the gripper. To prevent false diagnostic errors from being issued by the robotic system, an unambiguous labeling threshold was established: if the strawberry was captured in this boundary state (partially inside), it was still classified as ripe strawberries held in the gripper (Class 0). A sample was strictly labeled as an empty grasp (Class 1) if and only if the strawberry was completely absent from the effector. Similarly, a misgrasp (Class 2) was only triggered when the fruit was definitively confirmed as unripe.}

\textbf{SnapData} All images with resolution of 640$\times$480 from the miniature camera were captured during the snap-off stage. The dataset contained two classes: strawberry and background, where background referred to the scene outside the end-effector as seen by the camera when no strawberry was presented. The dataset was then split into training and validation sets in a 7:3 ratio, with 691 and 297 images, respectively. The inference results of SnapData were applied as input of LSTM classifier.

\begin{figure}
    \centering
    \includegraphics[width=0.5\linewidth]{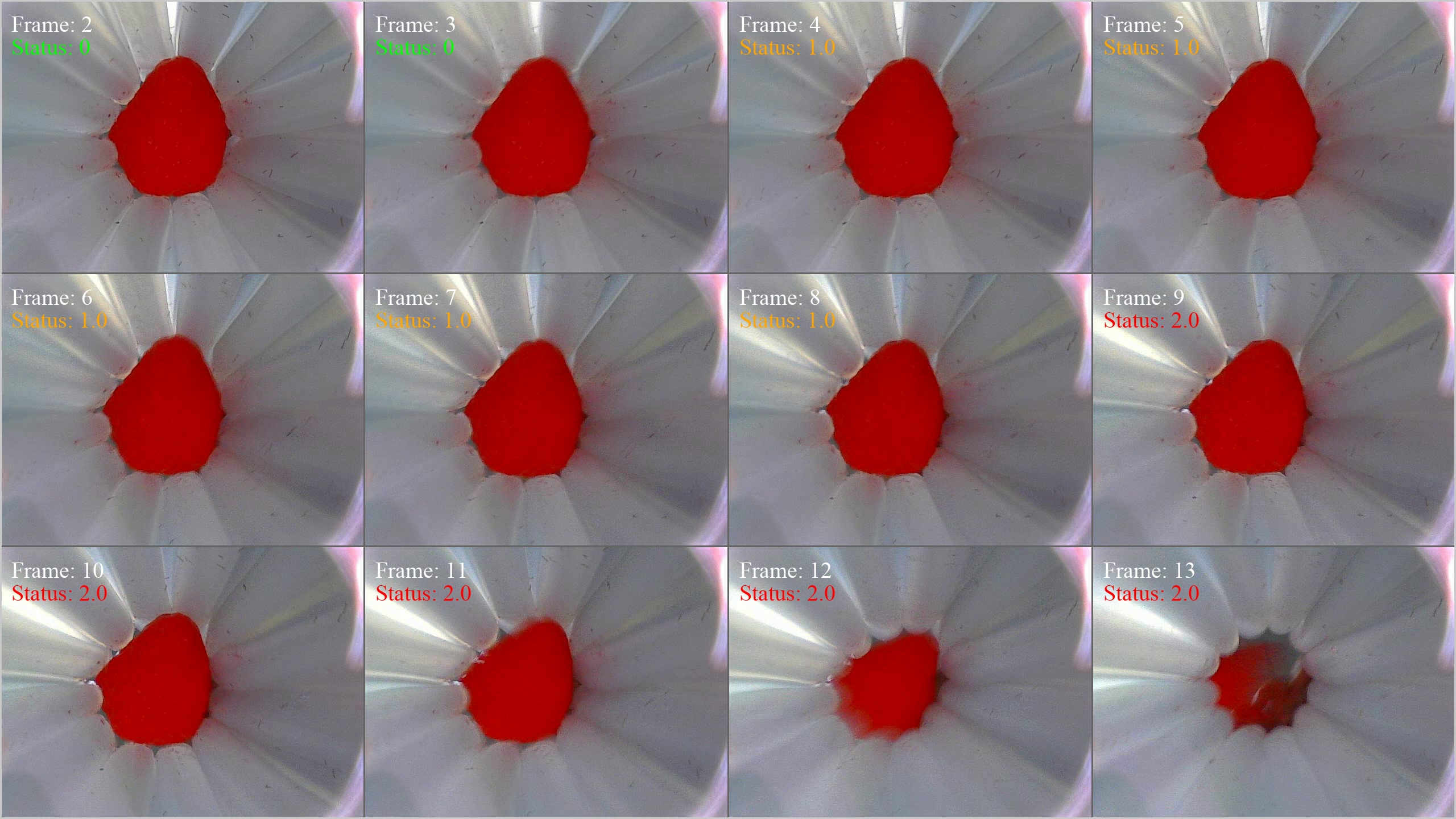}
    \includegraphics[width=0.42\linewidth]{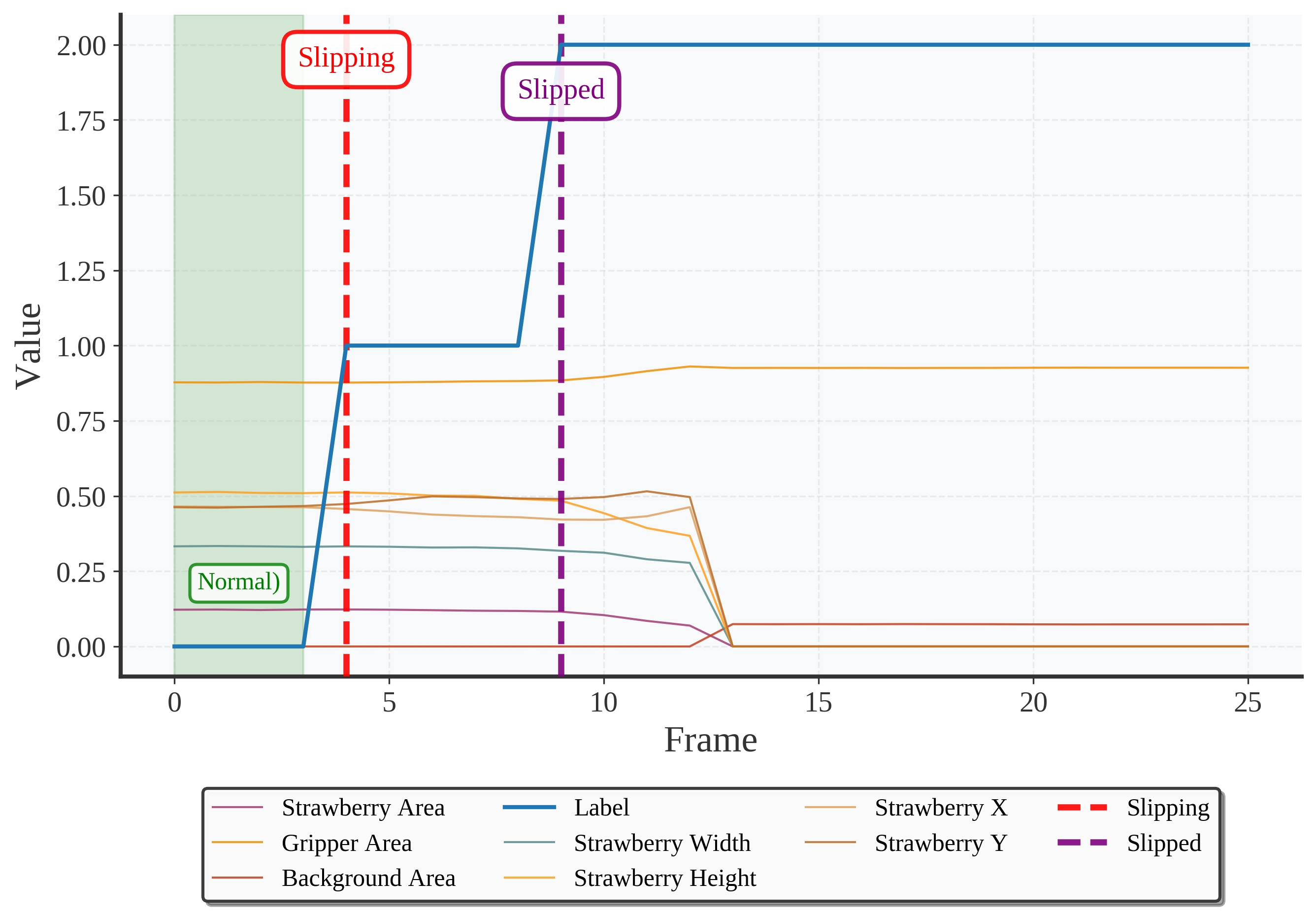}
    \caption{Strawberry slip process and status}
    \label{fig:snapdata}
\end{figure}

\textbf{SlipData} In each frame, key visual features—including the normalized strawberry area, normalized gripper area, normalized background area, and the width, height and center point of the strawberry—were extracted using a fine-tuned SRR-Net based on the SnapData. These features were structured in the format: $<$strawberry\_area$>$, $<$gripper\_area$>$, $<$background\_area$>$, $<w>$, $<h>$, $<x>$, $<y>$, $<$label$>$. This label indicated whether the strawberry was at risk of slipping from the gripper, with three possible statuses: normal (0), slipping (1), and slipped (2). A sliding window of 5 consecutive frames was used to form each input sequence. For each sequence, a label was assigned based on the next 3 frames immediately following the sequence. This structure enabled the training and validation of an  LSTM classifier to estimate the likelihood of strawberry slippage. The sample counts for labels 0, 1, and 2 were 719, 157, and 1962, respectively. Due to the brief duration of the slipping status during the snap-off stage, \textcolor{red}{a severe class imbalance occurred,} particularly for samples labeled as 1 (slipping). To mitigate the issue of sample imbalance, an oversampling method was introduced. Then, each category was split into training and validation sets at a 7:3 ratio.  
The images from frame 2 to frame 13 were presented, with the frame number and status displayed in the upper left corner of each subfigure in left of Fig. \ref{fig:snapdata}. On the right side were the variation curves of the seven feature components throughout the snap-off process, along with their corresponding labels. The horizontal axis represented the frame number, the vertical axis represented the feature values of the corresponding features, and the blue line indicated the labels for the corresponding frames. 

\section{Visual Fault Diagnosis and Self-Recovery Method}
\subsection{Overview}
% The HarvestFlex robot has successfully achieved strawberry harvesting in orchard environments. However, several challenges persisted that hindered harvesting performance during continuous operations. The core challenges included: (a) limited integration of visual perception; (b) grasping misalignment or failure resulting from positional offsets; and (c) reliance on human intervention to determine success or failure during the picking process, owing to the absence of automated evaluation and recovery mechanisms, as shown in Fig. \ref{fig-4}. %A video demo was available at \url{https://1drv.ms/v/c/e68b602748988e28/IQDbhbliJ1nPTqooSIfJg0S0AeDAfkL3JQGtFZfRcWEQImA?e=ISC7Mm}. 

\begin{figure*}%[]
  \centering
   \includegraphics[width=5in]{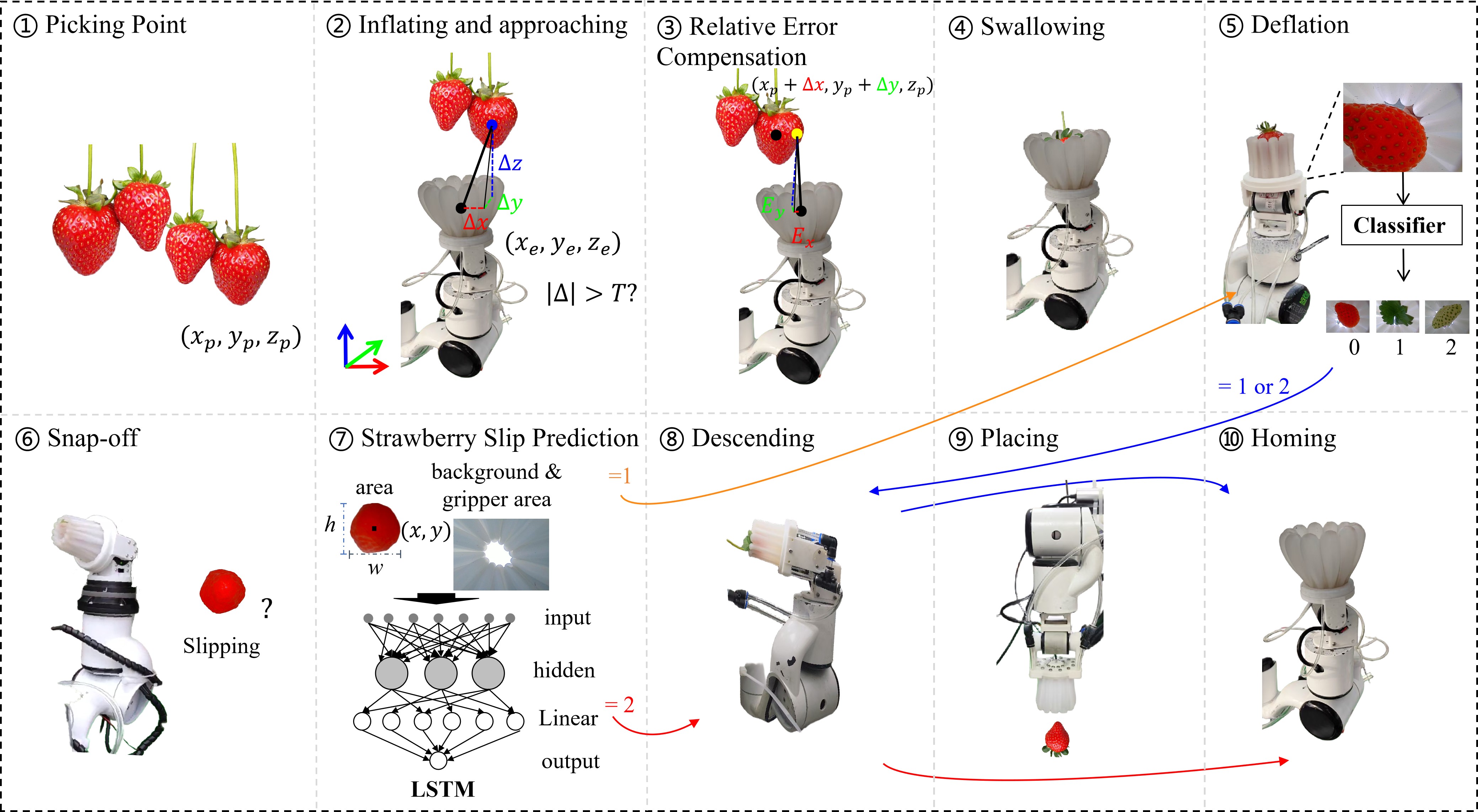}
   \caption{Harvesting fault and recovery in HarvestFlex}\label{fig-4}
\end{figure*}

To address the aforementioned issues, a vision-based harvest fault diagnosis and self-recovery architecture was developed \textcolor{red}{in Fig. \ref{fig-4}}. This architecture integrated three key components: an end-to-end multi-task perception network, a relative error compensation method, an early abort strategy based on the empty grasp/misgrasp adjustment and strawberry slip prediction. Specifically, the end-to-end multi-task perception network, SRR-Net, simultaneously performed detection, segmentation, and ripeness \textcolor{red}{regression}. A vision-based positional error compensation method was implemented to correct the $x$- and $y$-axis offsets between the picking point and the gripper center, thereby enhancing localization accuracy. To detect unsuccessful grasps, a micro-optical camera was embedded within the end-effector to monitor strawberry retention during the deflating stage and predict potential slippage risks during the snap-off stage. A MobileNet V3-Small classifier was applied to determine whether ripe strawberries \textcolor{red}{were} present at the end-effector during the deflating stage.  
An LSTM-based visual time-series prediction model provided real-time \textcolor{red}{slippage forecasts, enabling} a secondary harvesting attempt, the triggering of an early abort signal, or the continuation of subsequent actions. 

\subsection{End-to-End Multi-Task Perception Method: Detection, Segmentation and Ripeness \textcolor{red}{Regression}}
An end-to-end multi-task perception method, SRR-Net, was introduced to simultaneously perform detection, segmentation, and ripeness \textcolor{red}{regression} of both strawberries and gripper. This method integrated low- and high-level feature representations with ripeness and biological attributes to comprehensively perceive strawberries and reduce errors in ripeness regression. The method employed a lightweight network with shared weights. For strawberries, the perception task distinguished between ripe and unripe fruits by post-processing classification. For the robotic gripper, relative error between the gripper and the picking point was calculated.

\begin{figure*}%[]
  \centering
   \includegraphics[width=5in]{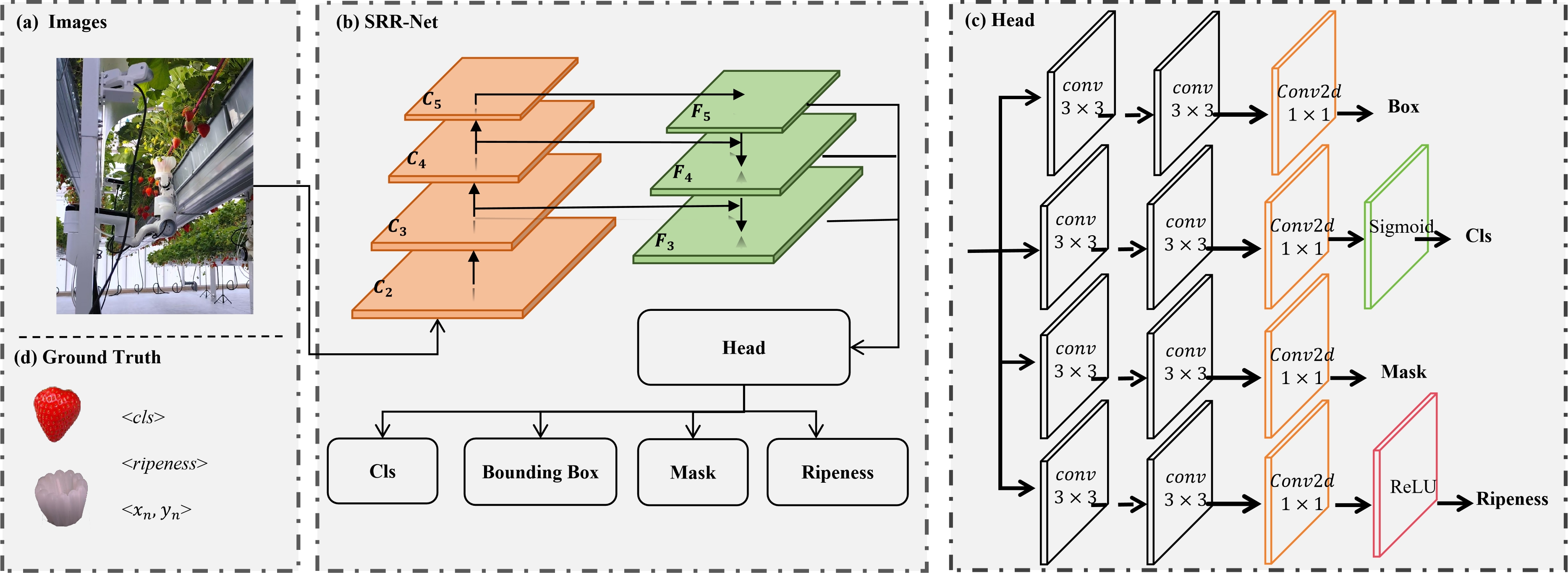}
    \caption{The architecture of end-to-end multi-task perception method}\label{fig-6}
\end{figure*}

This method was built on the general real-time detection and segmentation framework YOLOv11. The architecture consisted of a backbone, neck, and head: the backbone extracted features, the neck fused multi-scale features, and the head performed classification, bounding box regression, instance segmentation, and ripeness \textcolor{red}{regression}. Compared with the original YOLOv11, the primary modification was the addition of a parallel ripeness \textcolor{red}{regression} branch embedded within the head. \textcolor{red}{To supervise ripeness regression, the mean absolute error was used as the loss function.} An overview of the proposed end-to-end multi-task method was shown in Fig. \ref{fig-6}.

\subsection{Simultaneous Target-Gripper Detection for Relative Error Compensation} 
Traditional harvesting robots \textcolor{red}{relied} on calibration-based methods to transform the object's location from the camera frame to the robot arm frame.
% However, these approaches \textcolor{red}{were} susceptible to calibration errors and positioning inaccuracies in dynamic environments, exacerbated by system latency, making effective compensation challenging \citep{ge2023three}. 
\textcolor{blue}{However, these approaches were prone to positioning inaccuracies stemming from mechanical looseness and unpredictable fruit movement, making effective compensation challenging.}
In contrast, our proposed method detected the target strawberry and the gripper simultaneously in the same camera view, enabling direct estimation and correction of their relative error within that shared frame. This approach minimized the impact of absolute calibration inaccuracies and system latency, as the offset was computed relative to both elements in the same view, providing a more robust and real-time compensation strategy for improved picking accuracy in the HarvestFlex system.
\begin{figure}
    \centering
    \includegraphics[width=0.75\linewidth]{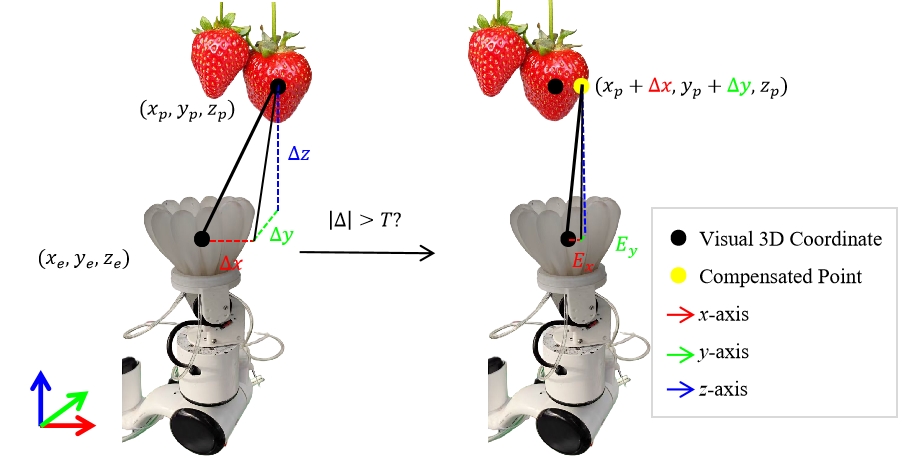}
    \caption{Simultaneous target-gripper detection for relative error compensation}
    \label{fig:positional_error}
\end{figure}

In the HarvestFlex arm’s workflow, a relative error compensation method was incorporated into the inflating and approaching and swallowing stages of the picking process, in Fig. \ref{fig:positional_error}. 
When the gripper approached the strawberry picking point, the 3D coordinates of both the end-effector and the picking point were obtained in the robot arm coordinate system. 
Specifically, using SRR-Net, the 2D image coordinates of the strawberry and the end-effector were extracted in the \textcolor{blue}{image} coordinate system. By combining these with the corresponding depth information, their 3D positions in the camera coordinate system were reconstructed and then transformed into the robot arm coordinate system through hand–eye \textcolor{blue}{coordinate transformation}. 
Let $(x_p,y_p,z_p)$ and $(x_e,y_e,z_e)$ denote the coordinates of the picking point and the end-effector in the robot arm frame, respectively. \textcolor{red}{The relative positional error ($\Delta x$, $\Delta y$, $\Delta z$) between the end-effector and the picking point was determined by the difference along each axis: $x_p - x_e$, $y_p - y_e$, and $z_p - z_e$.}

\textcolor{red}{During the harvesting process, HarvestFlex utilized a snap-off manner to pick strawberries. When operated by the robotic arm and end-effector, directly snap-off the target strawberry often created a pulling force between the fruit and the stem (peduncle), leading to abrasion and bruising at the contact area with the gripper. To prevent bruising or other mechanical damage to the grasped fruit during the snap-off action, the end-effector was intentionally elevated slightly above the target center during the swallowing phase. This upward offset mitigated the tensile force between the strawberry and its peduncle  \citep{11231358}. For this reason, only the $x$ and $y$ coordinate errors were considered for the estimation, while $z$-axis errors were ignored.}
\textcolor{red}{Given a maximum error tolerance threshold $T$,} when the end-effector reached a position beneath the picking point and the absolute error exceeded $T$, the compensated picking point coordinates $(x_c,y_c,z_c)$ were computed as:
\begin{equation}
x_c = 
\begin{cases}
x_p + k_x \Delta x & \text{if } |\Delta x| > T \\
x_p & \text{otherwise}
\end{cases}
\end{equation}
\begin{equation}
y_c = 
\begin{cases}
y_p + k_y \Delta y & \text{if } |\Delta y| > T \\
y_p & \text{otherwise}
\end{cases}
\end{equation}
\begin{equation}
z_c = z_p
\end{equation}
where,
% $T$ denoted the maximum tolerance of positional error for each axis, set to 10 mm. 
$k_x$ and $k_y$ refer to the scale factors of error. The robotic arm was then moved to the compensated point for strawberry grasping. \textcolor{red}{ The relative positional error compensation function was triggered only when the error along either the $x$-axis or the $y$-axis exceeded the maximum tolerance threshold. This condition was established because the system utilized a soft pneumatic end-effector, which could adaptively adjust its opening size to successfully grasp strawberries despite minor positional deviations. Accordingly, the maximum tolerable error along each coordinate axis was set to 10 mm. Furthermore, $k_x$ and $k_y$ were introduced as hyper-parameters to ensure that the robotic arm and end-effector arrived approximately or precisely below the target picking point after position compensation.} \textcolor{blue}{Based on empirical harvesting tests, these parameters were set to 1.0 and 0.5, respectively.}

\textcolor{red}{
For relative error compensation, the waiting time for the compensation signal was capped at 1.0 s.
When the visual perception system simultaneously recognized and tracked both the picking point and the end-effector, their relative error was calculated in real time via hand-eye coordinate transformation. If this error exceeded the maximum tolerance threshold within the 1.0 s allowable time window, a compensation signal was triggered to update the coordinates of the picking point. In cases where either the picking point or the gripper failed to be recognized or tracked, and this state persisted beyond the maximum timeout threshold, the system proceeded to execute the subsequent operations. Upon receiving the new picking point, the gripper re-approached the target before performing the subsequent actions. 
}

\subsection{Early Aborting Robotic Picking upon Grasp Adjustment and Slip Prediction}
\begin{figure*}[ht]
  \centering
  \begin{minipage}{0.35\linewidth}
    \centering
    \includegraphics[width=2in]{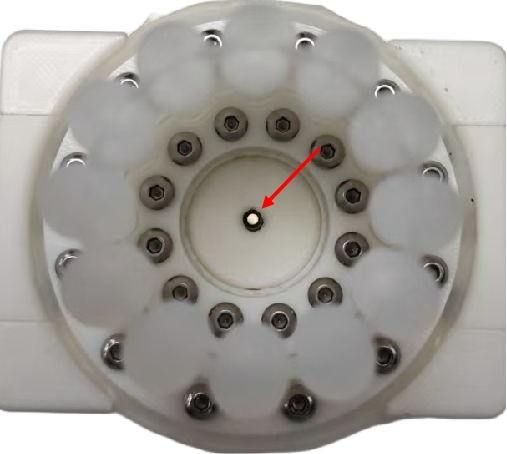}
  \end{minipage}
  \hfill
  \begin{minipage}{0.35\linewidth}
    \centering
    \includegraphics[width=1in]{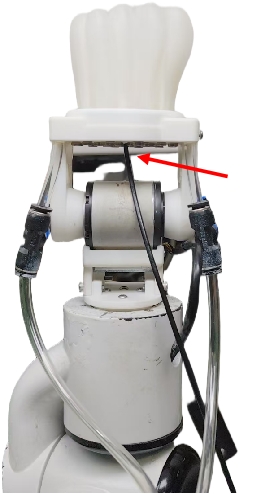}
  \end{minipage}
  \caption{Gripper with a micro-optical camera. (left) Top-view and (right) Side-view}
  \label{minicamera}
\end{figure*}

Empty grasp and misgrasp represented common failure modes in robotic strawberry harvesting, particularly during the deflating and snap-off stages. During the deflating stage, empty grasps occurred when the strawberry was not successfully enclosed by the end-effector, primarily due to localization errors or unexpected fruit motion. In the snap-off stage, strawberry slippage occurred when the picked strawberry unintentionally slipped, often due to a non-optimal end-effector pose or inadequate gripping force. To diagnose such faults effectively, a simple yet efficient method involved the installation of a miniature camera (JTS302, JiuTan, China) with a 3 mm diameter at the bottom of the end-effector, as shown in Fig. \ref{minicamera}. 
\textcolor{red}{A significant advantage of embedding a camera at the base of the end-effector was that it provided the most direct visual confirmation of whether a strawberry had been successfully harvested. Compared to mounting the camera on the robotic wrist, installing it inside the end-effector substantially reduced visual interference from adjacent strawberries, branches, and foliage surrounding the target picking point.} The micro-optical camera directly captured real-time images of the gripper area, enabling visual monitoring of strawberry presence. 

% Figure

\begin{figure}
    \centering
    \includegraphics[width=0.75\linewidth]{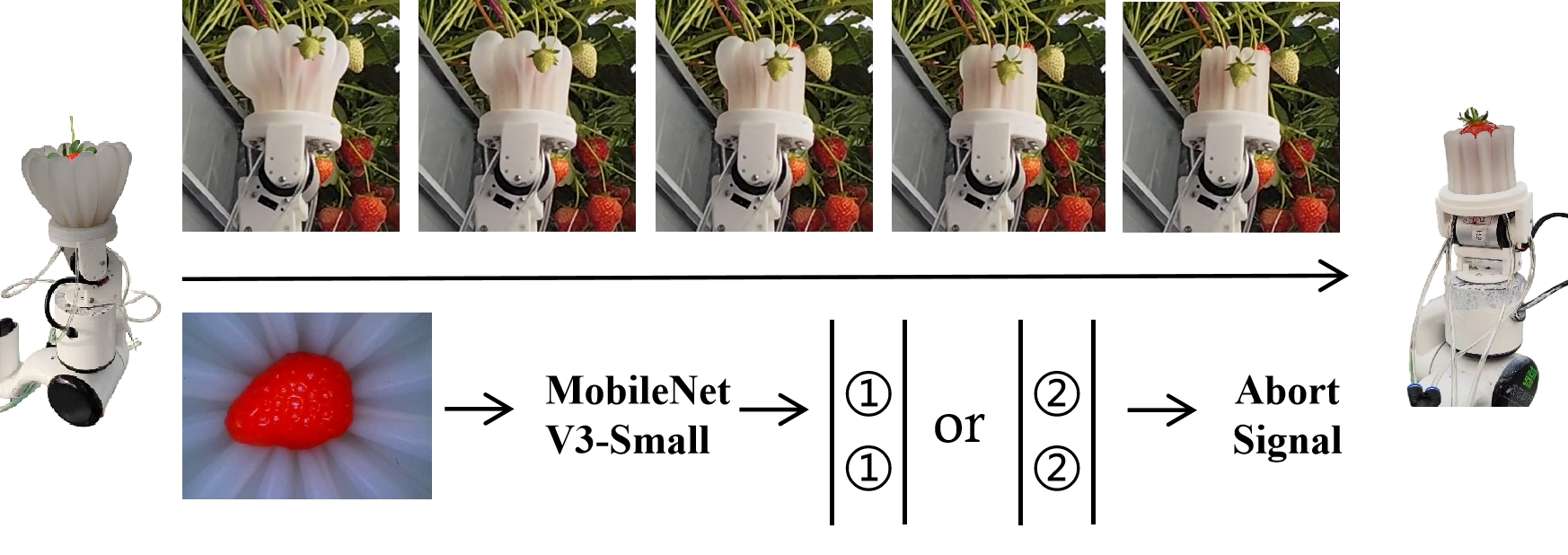}
    \caption{Grasp adjustment}
    \label{fig:empty_grasp}
\end{figure}

Once the deflating stage began, a MobileNet V3-Small  was used to determine whether a strawberry was present in the gripper, as shown in Fig. \ref{fig:empty_grasp}. If the classification result is 0 (strawberry correctly grasped), the robotic arm continued executing the subsequent picking actions. \textcolor{red}{If the classifier output the same error class (specifically, class 1 for an empty grasp or class 2 for a misgrasp of an unripe strawberry) for two consecutive frames, }the manipulator re-inflated the gripper, executed a descending motion, skipped the snapping and placing stages, and directly performed the homing motion to abort the current picking cycle. Instead, it immediately selected a new target strawberry and proceeded to the next picking cycle. The process flow, indicated by the blue line, was illustrated in Fig. \ref{fig-4}. 

\begin{figure}
\scriptsize
    \centering
    \includegraphics[width=0.95\linewidth]{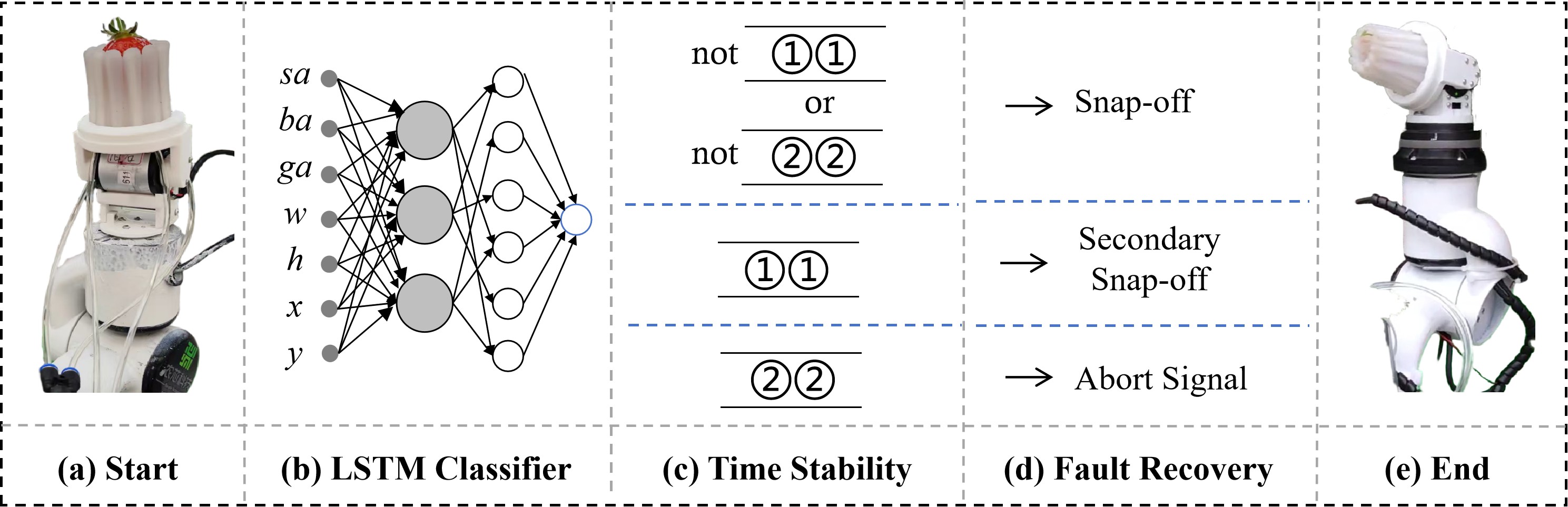}
    \caption{Strawberry slip prediction and self-recovery}
    \label{fig:lstm}
\end{figure}

For the slippage scenario, a time-series LSTM classifier was developed to predict whether the strawberry would slip from the end-effector during the snap-off stage in Fig. \ref{fig:lstm}. 
The LSTM classifier architecture was summarized in Table \ref{tab:lstm_cls}. The input layer of LSTM classifier had 7 features based a finetuned SRR-Net wtih strawberry and background classes. The LSTM layer consisted of 5 layers, a hidden size of 64, and a dropout of 0.2. This was followed by a Dropout layer with a rate of 0.3, then a Linear layer mapping from 64 to 3 units, and a Softmax layer applied along dimension 1. The output layer produced 3 classes. 
\begin{table}
\scriptsize
\caption{LSTM classifier architecture}
    \centering
    \begin{tabular}{l|c}
        \toprule
        Layer  &Number\\
        \midrule
        input  &7\\
        \multirow{3}{*}{LSTM} & layers=5\\
        & hidden=64\\
        & dropout=0.2\\
        Dropout&0.3\\
         Linear& (64, 3)\\
         Softmax& dim=1\\
        output  &3\\
        \bottomrule
    \end{tabular}
    \label{tab:lstm_cls}
\end{table}

The input of the LSTM classifier consisted of a sequence of features from the past 5 frames, represented as $<$ normalized strawberry area, normalized gripper area, normalized background area, $w$, $h$, $x$, $y >$. The normalized strawberry area, gripper area, and background area referred to the proportion of pixels in the image corresponding to the strawberry, the gripper, and the background, respectively. $x$,  $y$, $w$ and $h$ represented the center point, width and height of the strawberry. The model outputted the probability of slippage over the following three frames. 

\textcolor{red}{The outputs of the LSTM classifier were categorized into three states: a result of 0 meant the strawberry remained securely attached to the end-effector during snap-off stage (a successful pick); a result of 1 indicated the fruit was slipping from the gripper; and a result of 2 confirmed the strawberry had slipped completely. }To enhance reliability, a time-stability rule was implemented. If slipped was predicted in two consecutive frames, an early abort signal was triggered, the placing stage was skipped, and the next picking cycle began (as indicated by the red line in Fig. \ref{fig-4}). If two consecutive frames were slipping, the end-effector retracted, re-inflated, and re-performed the snap-off action (orange line in Fig. \ref{fig-4}). Otherwise, the robotic arm performed the subsequent actions. \textcolor{red}{It is worth noting that during the secondary snap-off process, the system exclusively reacted to the slipped state. Namely, upon predicting the slipped condition for two consecutive frames, the placing stage was bypassed; otherwise, the remaining steps were completed in sequence.}

\subsection{Paradigm of HarvestFlex Visual Fault Diagnosis and Self-Recovery}
\begin{figure}[ht]
\scriptsize
    \centering
    \includegraphics[width=1\linewidth]{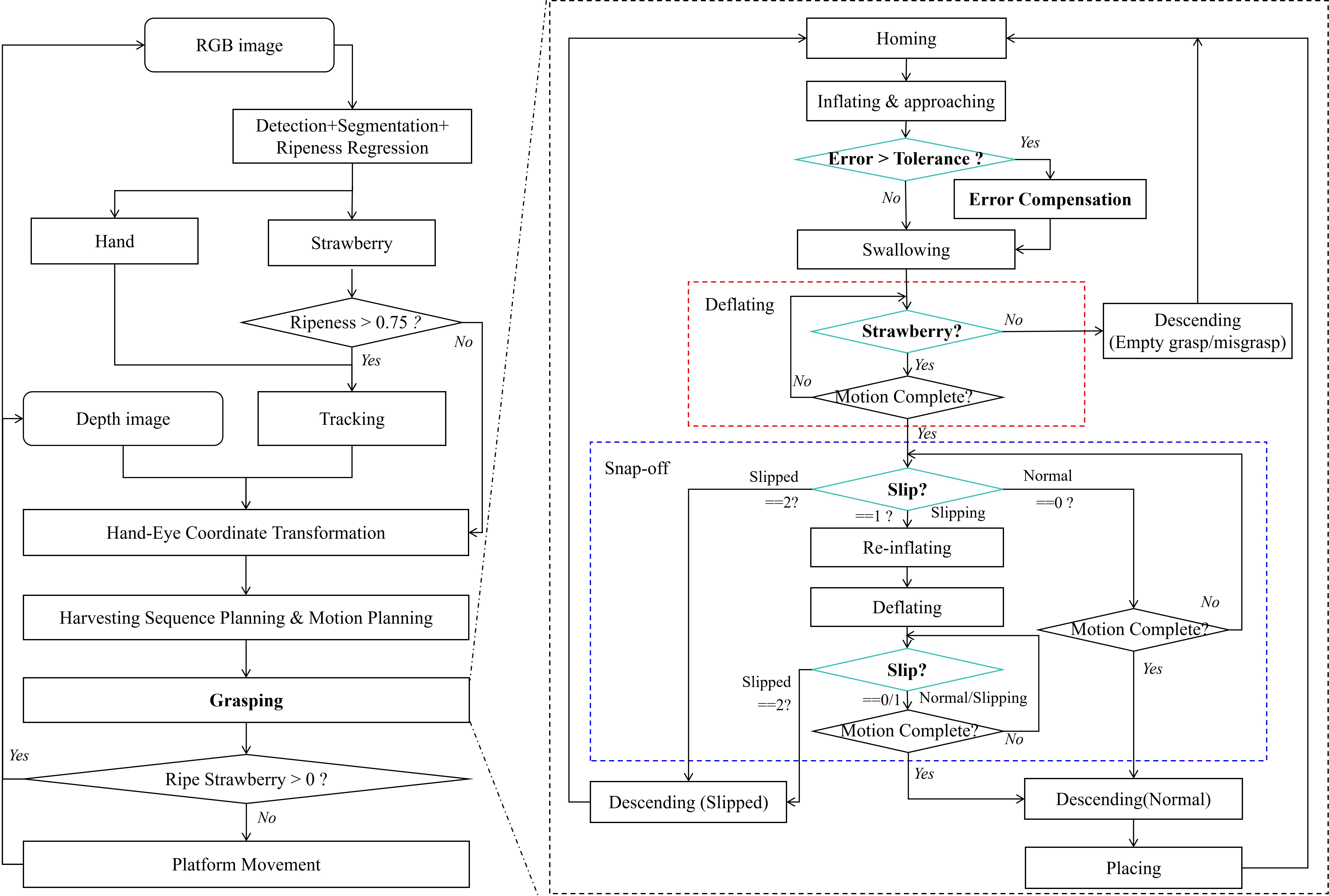}
    \caption{Paradigm of HarvestFlex visual fault diagnosis and self-recovery}
    \label{fig:paradigm}
\end{figure}

In HarvestFlex, visual fault diagnosis and self-recovery proceeded through the following steps, in Fig. \ref{fig:paradigm}. First, RGB and depth images were captured by a RealSense camera. The RGB images were processed by an end-to-end multi-task perception network to detect and segment strawberries and the gripper, and to estimate strawberry ripeness. Using a predefined ripeness threshold, fruits were classified as ripe or unripe, and only ripe strawberries were tracked in real time. Next, the camera coordinates of both tracked ripe and untracked unripe strawberries were transformed into the robot arm coordinate via hand-eye coordinate transformation. The end-effector’s coordinates, based on the camera coordinate system, were also extracted, tracked, and transformed to prevent track ID jumps. The system then planned the harvesting sequence for ripe strawberries and computed the robot arm motion trajectory to reach the (compensated) picking points. The gripper harvested each strawberry in sequence until no ripe fruits remained, at which point the mobile platform awaited new picking points. 

The relative positional error compensation method, incorporating visual feedback SRR-Net, and the early abort strategy based on MobileNet V3-Small and the LSTM-based strawberry slip prediction were implemented as follows. 
\textcolor{red}{Once the robotic arm received the coordinates for the picking point, }the gripper then inflated and approached a location directly beneath the target strawberry picking point. At this position, the \textcolor{red}{visual} relative error between the gripper and the strawberry position was calculated. If the error exceeded a predefined tolerance threshold, \textcolor{red}{the error compensation function was triggered and a compensated picking point was generated to correct the deviation before proceeding. Otherwise, the system bypassed the relative error compensation and directly executed the subsequent picking motion.}

Once aligned, the gripper executed swallowing and deflating motions to enclose the strawberry. During deflating stage, a micro-optical camera mounted on the gripper captured images, which were used to determine whether the strawberry was successfully gripped based on MobileNet V3-Small classifier. When no strawberry or an unripe strawberry was detected in the gripper for two consecutive frames, the robot triggered re-inflation, descending, and homing actions to begin a new picking cycle.
\textcolor{red}{ 
Upon completion of the deflating stage, the system executed the snap-off action to separate the stem from the strawberry. Concurrently during this phase, an LSTM classifier was employed to predict the strawberry's state within the end-effector over the subsequent three frames. Specifically, a prediction result of 0 indicated a normal separation, a result of 1 signified that the strawberry was slipping from the gripper, and a result of 2 denoted that the fruit had slipped.
}

\textcolor{red}{To enhance prediction stability, a time-stability rule was applied based on the three signal categories generated by the LSTM classifier: class 0 (normal), class 1 (slipping), and class 2 (slipped).}
When two consecutive slipped signals \textcolor{red}{(class 2)} were triggered, the robot aborted the snap-off stage, skipped the placing stage, and returned to its homing position. When two consecutive normal signals \textcolor{red}{(class 0)} were triggered, the robot continued the snap-off motion. 
Meanwhile, if two consecutive slipping signals \textcolor{red}{(class 1)} were triggered during the deflating and snap-off actions, the gripper was inflated and deflated to re-grasp the strawberry, and then performed a secondary snap-off, followed by the descending, placing, and homing steps. \textcolor{red}{Notably, during the secondary snap-off process, the system only responded to the slipped state. If this state was detected, the placing step was skipped, and the gripper proceeded directly to the descending and homing steps; otherwise, the subsequent actions were executed sequentially.}

\subsection{Implementation Details}
To evaluate the effectiveness of the proposed method, experiments were conducted on the FaultData, GraspData, SnapData, and SlipData benchmarks. The end-to-end multi-task perception framework was first trained and validated on the training and validation subsets of FaultData. 
\textcolor{red}{During this evaluation, the visual relative error between the strawberry and the gripper, positioned beneath the picking point, was measured. Additionally, the physical errors both before and after compensation were recorded to comprehensively assess picking accuracy. A MobileNet V3-Small classifier was trained and validated on the GraspData dataset. }
Furthermore, the end-to-end multi-task perception method was fine-tuned using the SnapData dataset. The outputs of the multi-task perception model were then applied to generate the SlipData for LSTM classifier. All models were trained using an NVIDIA RTX 4090 GPU. 
\textcolor{red}{The field experiments were also conducted at the Cuihu Farms. The visual fault diagnosis system was deployed and integrated into the HarvestFlex robot, which was equipped with an NVIDIA RTX 5000 Ada GPU.}

\section{Results}
\subsection{Evaluation Metrics}
For the detection and segmentation tasks, precision and recall were used to evaluate performance, following the definitions in \citep{10028728}. For strawberry ripeness \textcolor{red}{regression}, the mean absolute error (MAE) was adopted as the evaluation metric.
\textcolor{red}{For the relative error between the end-effector and the picked strawberry instance, the visual relative errors along the $x$- and $y$-axes under the robotic arm coordinate system were denoted as $\Delta x$, and $\Delta y$, respectively.}
Similarly, the corresponding physical errors along the $x$-, and $y$-axes were denoted as $\Delta x_w$ and $\Delta y_w$. The physical error after compensation, denoted as $E_x$ and $E_y$, was measured to validate the effectiveness of the relative error compensation method. Finally, \textcolor{blue}{MAE was also} used to evaluate the accuracy of the relative error and physical error before compensation and for each coordinate axis and can be formulated as:
\begin{equation}
  MAE = \frac{1}{n} \sum_{1}^{n} |e| , \text{ where } e = \{\Delta x, \Delta y, \Delta x_w, \Delta y_w, E_x, E_y\}
\label{eq:mae_equation}
\end{equation}
% In addition, FPS (frames per second) was used to evaluate the inference speed of the model.
\textcolor{red}{And the sample standard deviation (SD) was employed to quantify the variability of the errors.}

\subsection{Results of Detection, Segmentation, Ripeness \textcolor{red}{Regression}}
SRR-Net, which integrated multi-task learning for both strawberry and hand classes, was evaluated on the FaultData dataset for detection, segmentation, and ripeness regression in Table \ref{tbl-perception}. For bounding box prediction, SRR-Net achieved a precision of 0.90, recall of 0.81, mAP@50 of 0.88, and mAP@50-95 of 0.63 for strawberries, while for the hand class, the corresponding values were 0.97, 0.96, 0.98, and 0.79, respectively. In segmentation, SRR-Net reached a precision of 0.89, recall of 0.75, mAP@50 of 0.83, and mAP@50-95 of 0.45 for strawberries, and 0.97, 0.95, 0.96, and 0.66 for hands. In ripeness regression, SRR-Net achieved a mean absolute error of 0.04, highlighting its advantage in accurately assessing strawberry ripeness levels. 

Compared with YOLOv11 and YOLOv11-seg, SRR-Net demonstrated comparable detection and segmentation performance across both classes, with minor variations in individual metrics. \textcolor{red}{Overall, these results suggested that even when simultaneously performing detection, segmentation, and ripeness regression on both the strawberry and end-effector categories, SRR-Net maintained its performance on par with established baselines. Furthermore, it demonstrated solid capability in processing speed, proving to be an effective, well-balanced, end-to-end multi-task visual perception framework. }

\begin{table*}%[htpb]
\footnotesize
% \small
\scriptsize
\setlength{\tabcolsep}{1pt}
\centering
\caption{The detection, segmentation, and ripeness regression results on FaultData and and SnapData.}\label{tbl-perception}
\begin{tabular}{l|l|cccc|cccc|c}
\toprule
\multirow{2}{*}{Method} & \multirow{2}{*}{Class} & \multicolumn{4}{c|}{Box}  & \multicolumn{4}{c|}{Mask} & \multirow{2}{*}{MAE} \\
& & P & R & mAP@50  & mAP@50-95 & P & R & mAP@50 & mAP@50-95 & \\
\midrule
\textit{FaultData} & & & & & & & & & & \\
%\hline
\multirow{2}{*}{YOLOv11} & strawberry & 0.92 & 0.79 & 0.88 & 0.63 & - & - & - & - & - \\
& hand & 0.97 & 0.96 & 0.98 & 0.79 & - & - & - & - & -\\
%\hline
\multirow{2}{*}{YOLOv11-seg} & strawberry & 0.89 & 0.81 & 0.88 & 0.66 & 0.86 & 0.75 & 0.82 & 0.44 & - \\
& hand & 0.97 & 0.96 & 0.98 & 0.79 & 0.97 & 0.95 & 0.98 & 0.65 & - \\
\multirow{2}{*}{SRR-Net} & strawberry & 0.90 & 0.81 & 0.88 & 0.63 & 0.89 & 0.75 & 0.83 & 0.45 & 0.04 \\
& hand & 0.97 & 0.96 & 0.98 & 0.79 & 0.97 & 0.95 & 0.96 & 0.66 & -  \\
\bottomrule
\textit{SnapData} & & & & & & & & & \\
\multirow{2}{*}{SRR-Net} & strawberry & 0.99 & 0.99 & 0.99 & 0.98 & 0.99 & 0.99 & 0.99 & 0.98 & -  \\
& background & 0.93 & 0.91 & 0.97 & 0.86 & 0.93 & 0.91 & 0.96 & 0.86 & - \\
\bottomrule
\end{tabular}
\end{table*}

\begin{figure}[ht]
\scriptsize
    \centering
    % 第一行
    \begin{minipage}{0.31\textwidth}
        \centering
        \includegraphics[width=\linewidth]{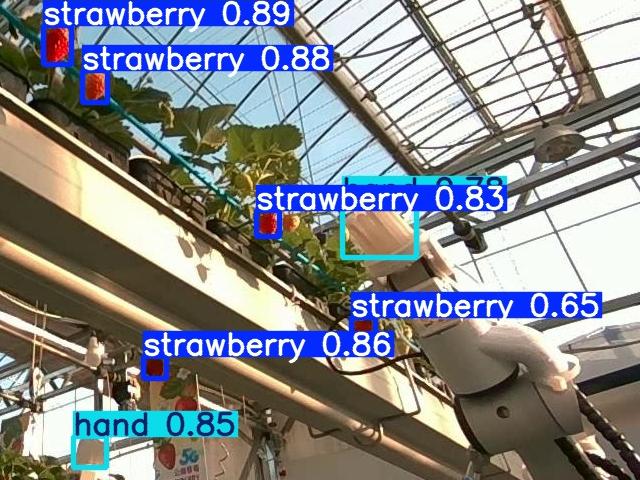}
    \end{minipage}
    \hfill
    \begin{minipage}{0.31\textwidth}
        \centering
        \includegraphics[width=\linewidth]{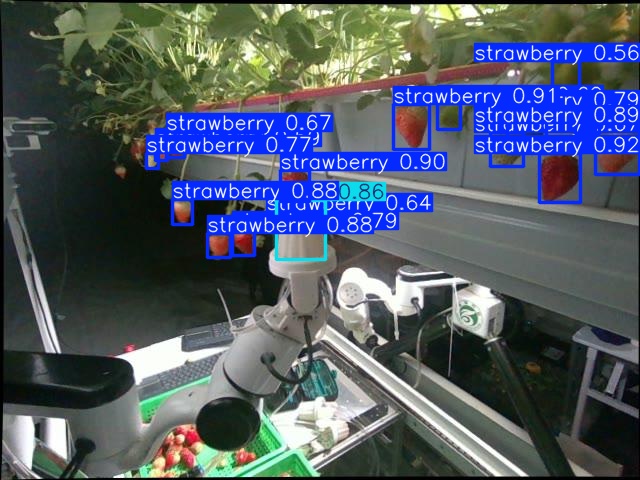}
    \end{minipage}
    \hfill
    \begin{minipage}{0.31\textwidth}
        \centering
        \includegraphics[width=\linewidth]{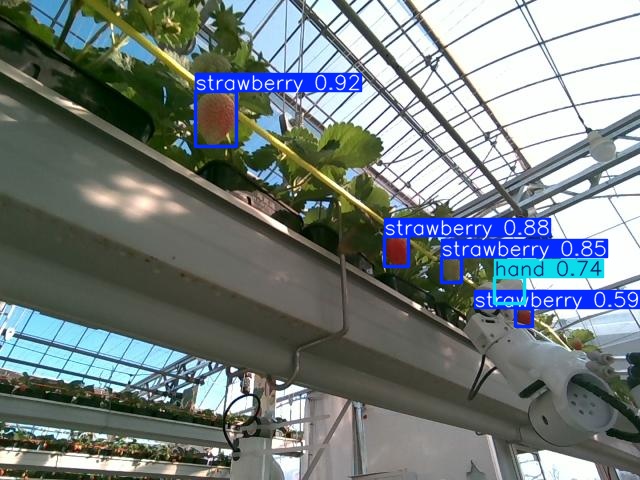}
    \end{minipage}

    \vspace{0.2cm}
    (a) YOLOv11
    \vspace{0.2cm}
    
    % 第二行
    \begin{minipage}{0.31\textwidth}
        \centering
        \includegraphics[width=\linewidth]{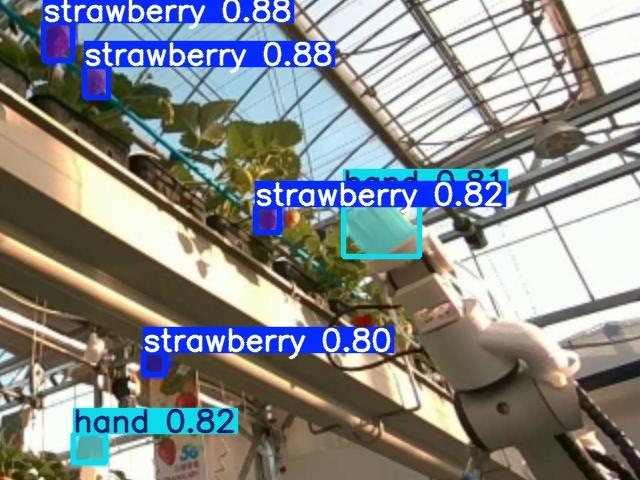}
    \end{minipage}
    \hfill
    \begin{minipage}{0.31\textwidth}
        \centering
        \includegraphics[width=\linewidth]{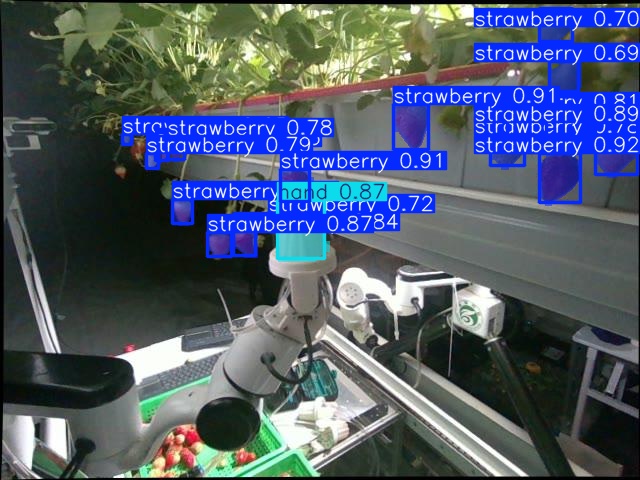}
    \end{minipage}
    \hfill
    \begin{minipage}{0.31\textwidth}
        \centering
        \includegraphics[width=\linewidth]{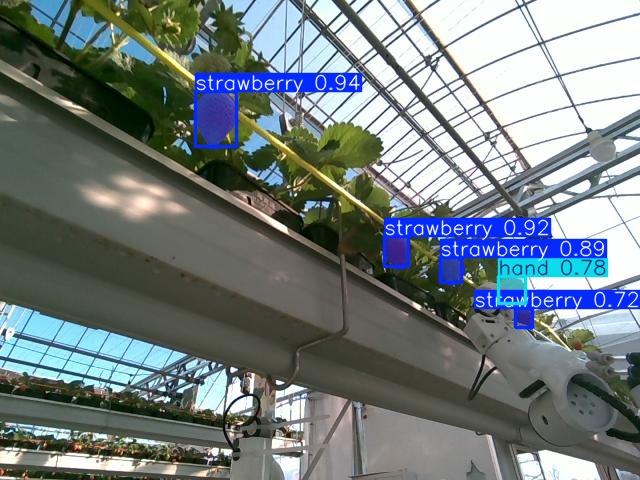}
    \end{minipage}
    
    % \vspace{0.5cm}
    \vspace{0.2cm}
    (b) YOLOv11-seg
    \vspace{0.2cm}
    
    % 第三行
    \begin{minipage}{0.31\textwidth}
        \centering
        \includegraphics[width=\linewidth]{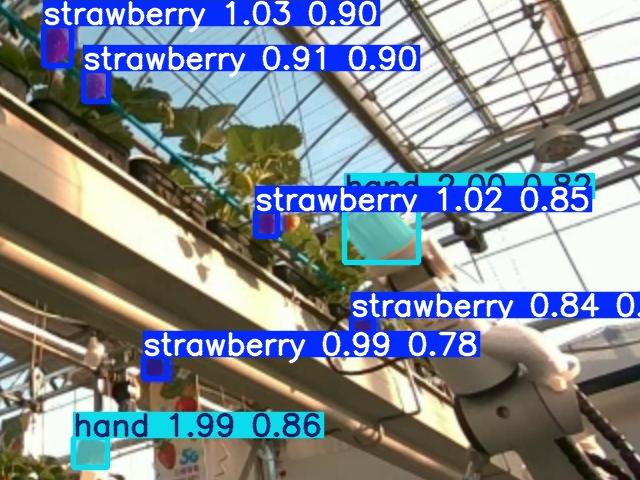}
    \end{minipage}
    \hfill
    \begin{minipage}{0.31\textwidth}
        \centering
        \includegraphics[width=\linewidth]{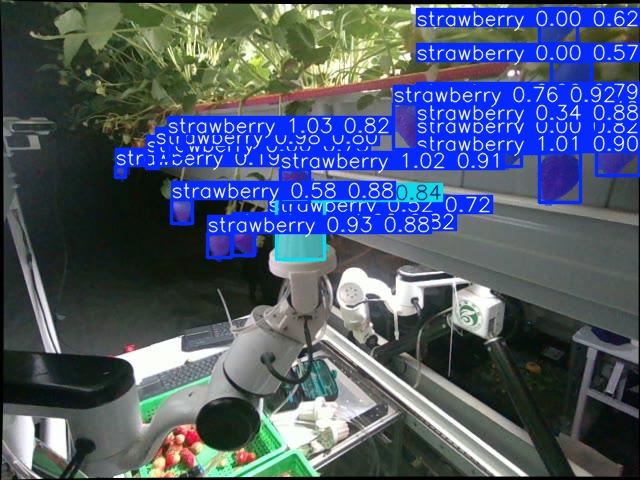}
    \end{minipage}
    \hfill
    \begin{minipage}{0.31\textwidth}
        \centering
        \includegraphics[width=\linewidth]{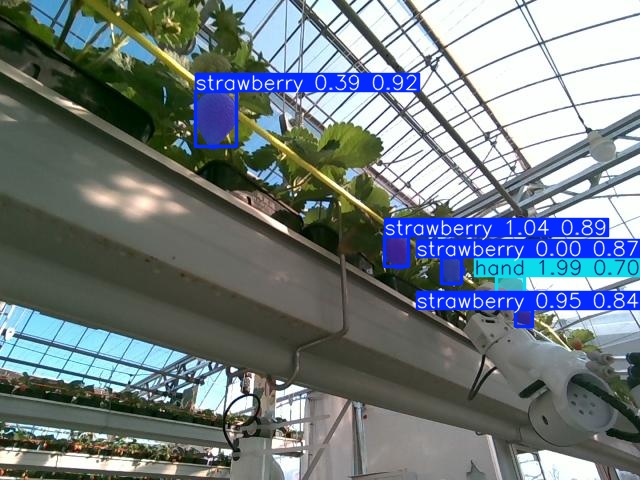}
    \end{minipage}
    
    \vspace{0.2cm}
    (c) SRR-Net
    \caption{Visualization results}
    \label{fig:visual}
\end{figure}

To more intuitively observe the performance of SRR-Net, the visualization results of SRR-Net, YOLOv11, and YOLOv11-seg on the FaultData dataset are presented in Fig. \ref{fig:visual}. In Fig. \ref{fig:visual} (a) and (b), strawberries and end-effectors were represented with bounding boxes, with the target class label and confidence score displayed above each box. The confidence threshold was set to 0.5. The ripeness value of SRR-Net was displayed between the target class label and the confidence score.
%Overall, the visual comparisons illustrated that SRR-Net delivered superior performance compared with YOLOv11 and YOLOv11-seg. 
\textcolor{blue}{Overall, the visual comparisons showed that SRR-Net performed at a similar level to YOLOv11 and YOLOv11-seg. }

\subsection{Evaluation of Relative Error Compensation}
To validate the effectiveness of relative error compensation, the 3D coordinates of the strawberry picking point ($x_s$,  $y_s$, $z_s$), the gripper positioned beneath the picking point ($x_e$, $y_e$, $z_e$), the visual relative error before compensation ($\Delta x$, $\Delta y$), the \textcolor{red}{physical} error before compensation ($\Delta x_w$, $\Delta y_w $) , the compensated picking point ($x_{ce}$,  $y_{ce}$, $z_{ce}$), and the \textcolor{red}{physical} relative error after compensation ($E_x$, $E_y$) in the robot arm coordinate system were measured in Table \ref{tbl2}. 
% Due to the requirement of snap-off, the rise distance in the swallowing stage exceeded the height of the strawberry. To simplify the experiments and reduce computational overhead, the relative error along the $z$-axis was not considered.

From Table \ref{tbl2}, the visual and \textcolor{red}{physical} errors before compensation were compared, and the \textcolor{red}{physical} errors after compensation were recorded. For example, in the first row, the visual relative errors on the $x$- and $y$- axis  were 23 mm and -4 mm, while the corresponding ground-truth errors were 17 mm and 4 mm. The absolute differences between visual and ground-truth errors were therefore 6 mm and 8 mm. After compensation, the physical errors of the $x$- and $y$- axis were reduced to 2 mm and 6 mm, both below the defined threshold.

% Before compensation, the mean \textcolor{red}{absolute} errors (MAE) $\pm$ standard deviation (SD) were 14.07 $\pm$ 6.91 mm on the $x$-axis and 8.64 $\pm$ 5.56 mm on the $y$-axis, while the mean physical errors $\pm$ standard deviation were 11.52 $\pm$ 4.49 mm and 5.15 $\pm$ 6.16 mm, respectively—indicating that visual estimation tended to overestimate the actual physical errors. 
\textcolor{blue}{Before compensation, the visual mean absolute errors (MAE $\pm$ SD) were 14.40 $\pm$ 7.12 mm and 9.00 $\pm$ 5.42 mm on the $x$- and $y$-axes, respectively. In contrast, the physical MAE $\pm$ SD were 11.50 $\pm$ 4.42 mm and 5.25 $\pm$ 6.25 mm, indicating that visual estimation tended to overestimate the actual physical errors.}

% With the proposed compensation method, the mean absolute physical error between the strawberry point and the gripper were further reduced to 3.12 $\pm$ 4.00 mm ($x$-axis) and 4.11  $\pm$ 4.38 mm ($y$-axis). 
\textcolor{blue}{With the proposed compensation method, the physical MAE $\pm$ SD between the target center and the gripper was further reduced to 3.12 $\pm$ 3.94 mm on the $x$-axis and 4.06 $\pm$ 4.33 mm on the $y$-axis.} The slightly larger residual $y$-axis error was likely due to the difficulty in achieving high-precision motor control over very small movement ranges. Despite this phenomenon, the relative error compensation method demonstrated a clear advantage in aligning the gripper with the target and in reducing grasping errors.

\begin{table}[ht]
\caption{\textcolor{blue}{Relative errors between the end-effector and the picking point in the arm coordinate system (unit: mm). $(x_s, y_s, z_s)$ and $(x_e, y_e, z_e)$ represented the initial picking point and end-effector coordinates, respectively. Their visual and physical relative errors were denoted by $(\Delta x, \Delta y)$ and $(\Delta x_w, \Delta y_w)$. After compensation, the updated picking point became $(x_{ce}, y_{ce}, z_{ce})$, yielding a physical relative error of $(E_x, E_y)$. The final two rows show the mean absolute error MAE and standard deviation SD.}}
\scriptsize
\label{tbl2}
\centering
\begin{tabular}{@{}c|c|c|c|c|c@{}}
\toprule
$(x_s, y_s, z_s)$ & $(x_e, y_e, z_e)$ & $(\Delta x$, $\Delta y)$ & $(\Delta x_w$, $\Delta y_w) $ & $(x_{ce}$, $y_{ce}$, $z_{ce})$ & $(E_x$, $E_y$) \\
\midrule
$(709, 221, 706)$ & $(686, 225, 647)$ &	$(23,  -4)$	& $(17,   4)$ & $(732, 219, 706)$ & $(2, 6)$\\
$(464, 232, 710)$ & $(450, 249, 652)$ & $(14, -17)$ & $( 8, -13)$ & $(478, 224, 710)$ & $(5, -8)$\\
$(377, 245, 693)$ & $(362, 255, 626)$ & $(15, -10)$ & $( 7,  -3)$ & $(392, 240, 693)$ & $(3, 4)$\\
$(700, 222, 699)$ & $(681, 224, 638)$ & $(19,  -2)$ & $(16,   3)$ & $(719, 221, 699)$ & $(0, 3)$\\
$(532, 234, 710)$ & $(518, 245, 649)$ & $(14, -11)$ & $(11,  -4)$ & $(546, 229, 710)$ & $(3, -3)$\\
$(711, 244, 711)$ & $(692, 249, 651)$ & $(19,  -5)$ & $( 9,  -5)$ & $(730, 242, 711)$ & $(0, 2)$\\
$(320, 244, 692)$ & $(316, 256, 628)$ & $( 4, -12)$ & $( 4,  -3)$ & $(324, 238, 692)$ & $(0, 2)$\\
$(468, 236, 703)$ & $(460, 245, 653)$ & $( 8,  -9)$ & $(13,  -1)$ & - &  -\\
$(652, 235, 712)$ & $(631, 239, 651)$ & $(21,  -4)$ & $(20,   6)$ & $(673, 233, 712)$ & $(-4, -2)$\\
$(816, 220, 699)$ & $(793, 223, 642)$ & $(23,  -3)$ & $(17,   7)$ & $(839, 219, 699)$ & $(-6,  5)$\\
$(393, 238, 706)$ & $(386, 245, 643)$ & $( 7,  -7)$ & $( 7,   3)$ & -  & -\\
$(734, 222, 699)$ & $(712, 224, 638)$ & $(22,  -2)$ & $(12,   5)$ & $(756, 221, 699)$ & $(-5,  4)$\\
$(445, 230, 711)$ & $(436, 249, 646)$ & $( 9, -19)$ & $(10, -10)$ & $(454, 221, 711)$ & $(-4,  6)$\\
$(629, 234, 721)$ & $(611, 244, 659)$ & $(18, -10)$ & $(18,  -3)$ & $(647, 229, 721)$ &  $(6,  6)$\\
$(299, 246, 693)$ & $(295, 259, 631)$ & $(4,  -13)$ & $(10,  -8)$ & $(303, 240, 693)$ & $(-2,  1)$\\ 
$(453, 233, 701)$ & $(443, 251, 637)$ & $(10, -18)$ & $(10,  -8)$ & $(463, 224, 701)$ & $(7,  -1 )$\\
$(752, 213, 708)$ & $(729, 218, 645)$ & $(23,  -5)$ & $(13,   7)$ & $(775, 211, 708)$ & $(0,  4)$\\
$(667, 215, 708)$ & $(645, 222, 645)$ & $(22,  -7)$ & $(14,   0)$ & $(689, 212, 708)$ & $(-4, -6)$\\
$(307, 246, 693)$ & $(305, 252, 633)$ & $( 2,  -6)$ & $( 6,   0)$ & - & - \\
$(467, 232, 700)$ & $(456, 248, 634)$ & $(11, -16)$ & $( 8, -12)$ & $(478, 224, 700)$ & $(-2 , 6)$ \\
\midrule
\multicolumn{2}{c|}{Mean absolute error MAE} & $(14.40, 9.00)$ & $(11.50, 5.25 )$ & - &  $(3.12 , 4.06 )$ \\
\multicolumn{2}{c|}{Standard deviation SD} & $( 7.12,  5.42)$& $(4.42, 6.25)$ & - &  $(3.94, 4.33)$ \\
\bottomrule
\end{tabular}
\end{table}

\subsection{Evaluation of Empty Grasp or Misgrasp}
MobileNet-V3 Small was trained and validated based on the GraspData. The classification results were presented in Table \ref{tab:mobilenet-result}. Class 0, 1, and 2 \textcolor{red}{represented} the presence of strawberries, the without strawberry, and the unripe strawberry in the end-effector, respectively. 
From Table \ref{tab:mobilenet-result}, it can be observed that the precision, recall, and F1-score for all classes \textcolor{red}{were} all 1.00, indicating that the model \textcolor{red}{achieved} optimal classification performance for each category and closely \textcolor{red}{matched} the ground-truth labels. \textcolor{red}{In the grasp evaluation process, the most critical and challenging aspect was accurately distinguishing whether the strawberry was located inside or outside the end-effector during the early stages of the deflating phase. For example, scenarios where the fruit was situated directly above the end-effector or where only the bottom of the strawberry was grasped led to model misclassifications. Such inaccuracies resulted in incorrect control responses from the robotic arm, thereby disrupting the normal operation of the harvesting robot. To prevent these misjudgments, an empty grasp was identified exclusively when a strawberry remained completely outside the end-effector. For strawberries successfully enveloped by the gripper, the grasping status was categorized as either a normal grasp or a misgrasp based on the strawberry ripeness. Because of these stringent and clear-cut class boundaries, the data distribution possessed highly distinct discriminative standards. Consequently, after feature extraction by the model, it achieved a perfect classification performance of 1.0.}

\begin{table} [ht]
\scriptsize 
    \centering
    \caption{Results of empty grasp or misgrasp. \textcolor{red}{Class 0, 1, and 2 \textcolor{red}{represented} the presence of strawberries, the without strawberry, and the unripe strawberry in the end-effector, respectively}}
    \begin{tabular}{c|ccc}
    \toprule
         Classes&  Precision&  Recall&F1-score\\
    \midrule
         0&  1.00&  1.00&1.00\\
         1&  1.00&  1.00&1.00\\
 2& 1.00& 1.00&1.00\\
 \bottomrule
    \end{tabular}
    \label{tab:mobilenet-result}
\end{table}

\begin{figure}[h]
\scriptsize
    \centering
    \includegraphics[width=0.32\linewidth]{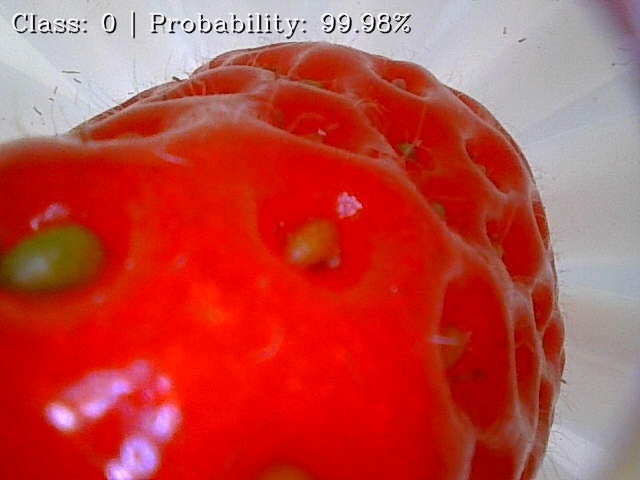}
    \includegraphics[width=0.32\linewidth]{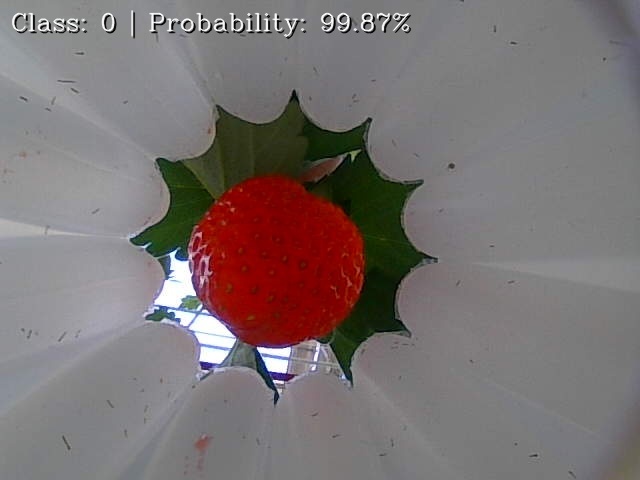}
    \includegraphics[width=0.32\linewidth]{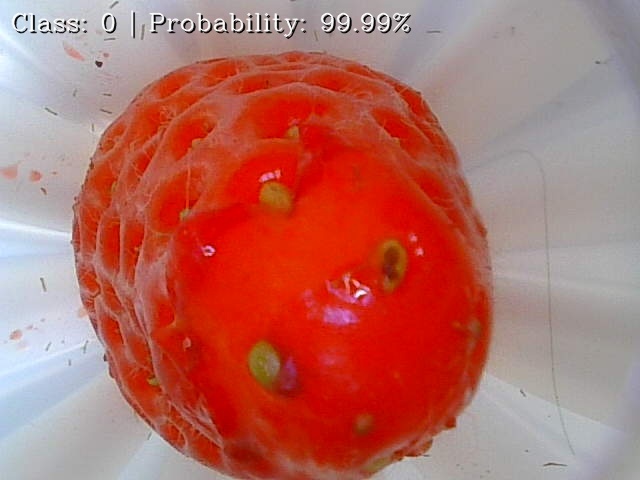}
    \vspace{0.2cm}
    (a) Strawberry in the end-effector\\
    % \vspace{0.05cm}
    \includegraphics[width=0.32\linewidth]{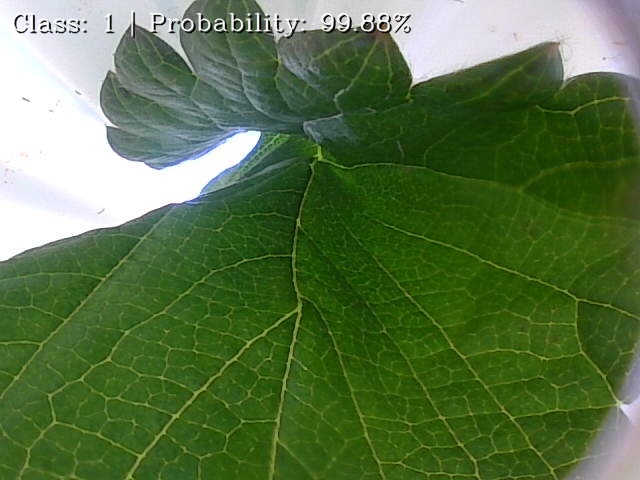}
    \includegraphics[width=0.32\linewidth]{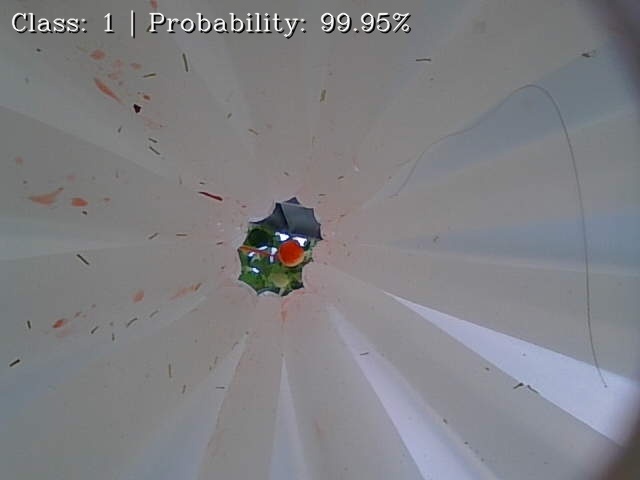}
    \includegraphics[width=0.32\linewidth]{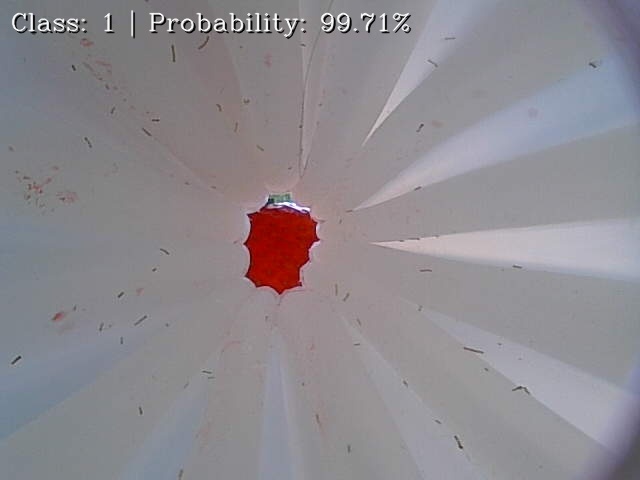}
    \vspace{0.2cm}
    (b) No strawberry in the end-effector\\
    % \vspace{0.2cm}
    \includegraphics[width=0.32\linewidth]{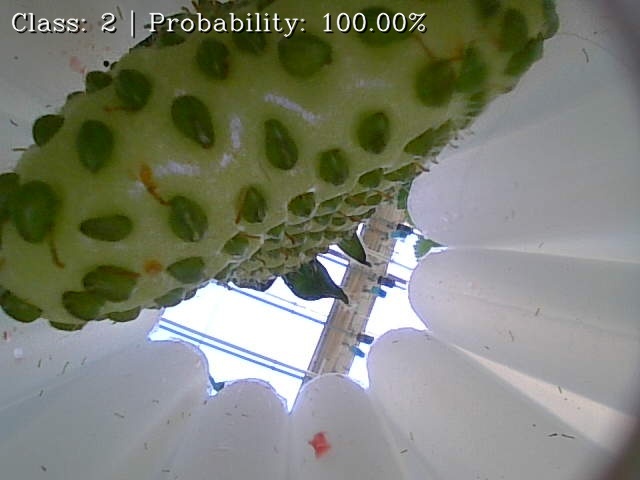}
    \includegraphics[width=0.32\linewidth]{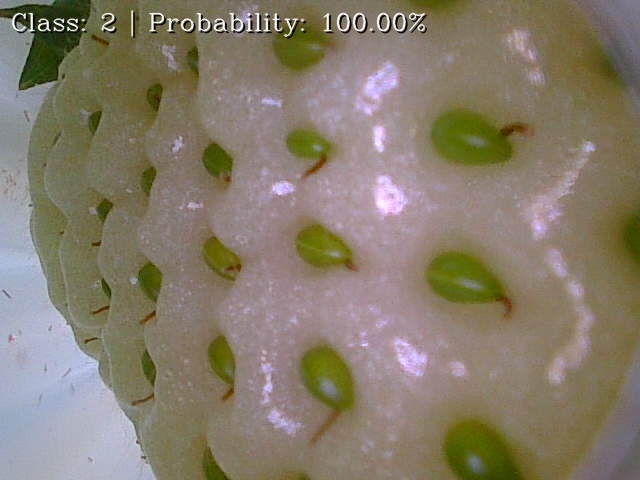}
    \includegraphics[width=0.32\linewidth]{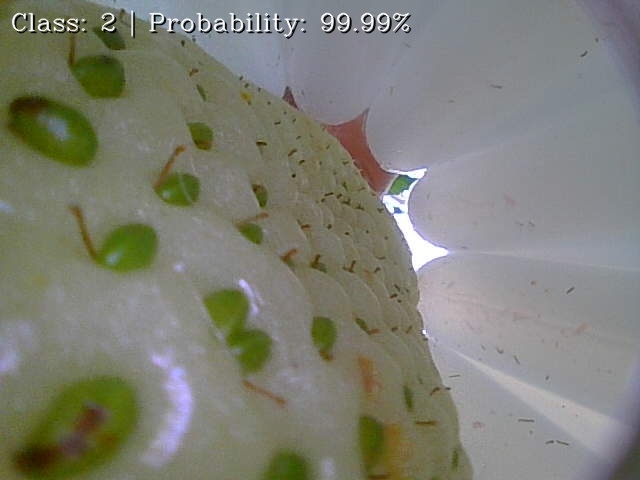}
    \vspace{0.2cm}
    (c) Misgrasp of an unripe strawberry
    % \vspace{0.2cm}
    \caption{Visualization of successful grasp, empty grasp or misgrasp}
    \label{fig:visulizationmobilenet}
\end{figure}

To observe the classification results of MobileNet-V3 Small, the classification outputs of some images were visualized in Fig. \ref{fig:visulizationmobilenet}. The upper-left corner of each image was annotated with the predicted class and the corresponding probability. From Fig. \ref{fig:visulizationmobilenet}, it can be concluded that MobileNet-V3 Small achieved good classification performance.

Based on the classification results of MobileNet-V3 Small, different responses were triggered on the robotic arm. When the classification result was 0, the robotic arm continued to execute the subsequent operations. \textcolor{red}{When two consecutive frames were identically classified as either 1 or 2}, the end-effector performed secondary inflation to prevent damage to the strawberry, then skipped the snap-off and placing stages and directly moved the end-effector downward to a position below the strawberry, thereby ending the current picking cycle.

\subsection{Evaluation of Slip prediciton}
Before performing LSTM classification in the snap-off stage, SRR-Net was first fine-tuned based on SnapData. These images contained two classes---strawberry and background---which were used to count the number of pixels belonging to the strawberry and background regions during the snap-off stage. The results of SRR-Net on SnapData were presented in Table \ref{tbl-perception}. The precision and recall of the segmentation task reached 0.99 and 0.99 for strawberries, and 0.93 and 0.91 for the background, respectively, indicating that the model achieves highly accurate segmentation performance. 

\begin{table}
\scriptsize
    \centering
    \caption{LSTM classifier result on the SlipData. \textcolor{red}{class 0, 1 and 2 represented strawberries remaining in the gripper, slipping and slipped, respectively. }}
    \begin{tabular}{c|ccc}
        \toprule
          Class&Precision&  Recall& F1-socre\\
          \midrule
          0&0.94&  0.89& 0.92\\
          1&0.86&  0.94& 0.90\\
          2& 0.99& 0.94 &0.96\\
         \bottomrule
    \end{tabular}
    \label{tab:lstm_slip_res}
\end{table}
The normalized strawberry area, normalized background area, normalized hand area, width and height of strawberry, center point of strawberry as input of LSTM classifier to train and validate using SlipData. The results of the LSTM classifier on the validation set of SlipData were presented in Table \ref{tab:lstm_slip_res}, where class 0, 1 and 2 represented strawberries remaining in the gripper, slipping and slipped, respectively. As shown in Table \ref{tab:lstm_slip_res}, the LSTM classifier achieved high accuracy despite the imbalance among class samples. 
\textcolor{red}{After resampling, the F1-scores for Classes 0, 1, and 2 were 0.92, 0.90, and 0.96, respectively, indicating that the model effectively predicted whether strawberry slippage would occur over the subsequent three frames. A prominent feature of an slipping strawberry was the variation in its distance from the bottom camera. Visually, this translated to decreases or fluctuations in the pixel areas of the end-effector, the background, and the strawberry itself, along with shifts in the strawberry's center point coordinates. Furthermore, instances of a completely slipped strawberry were typically characterized by image content dominated by the end-effector region, coupled with a diminished strawberry pixel area and an expanded background region. }

\begin{figure*}[ht]
\scriptsize
  \centering
  \includegraphics[width=3.1in]{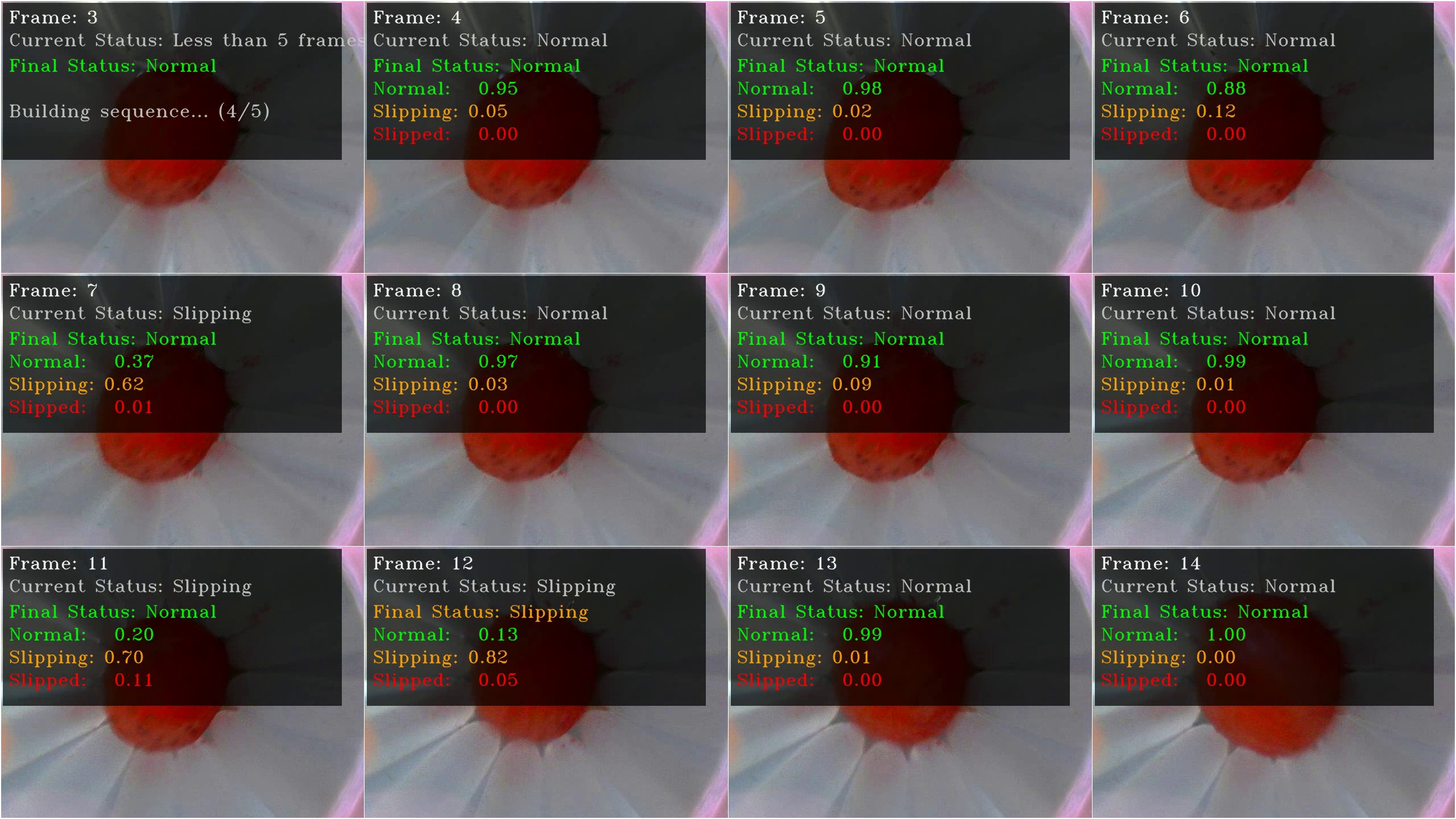}
  \includegraphics[width=2.2in]{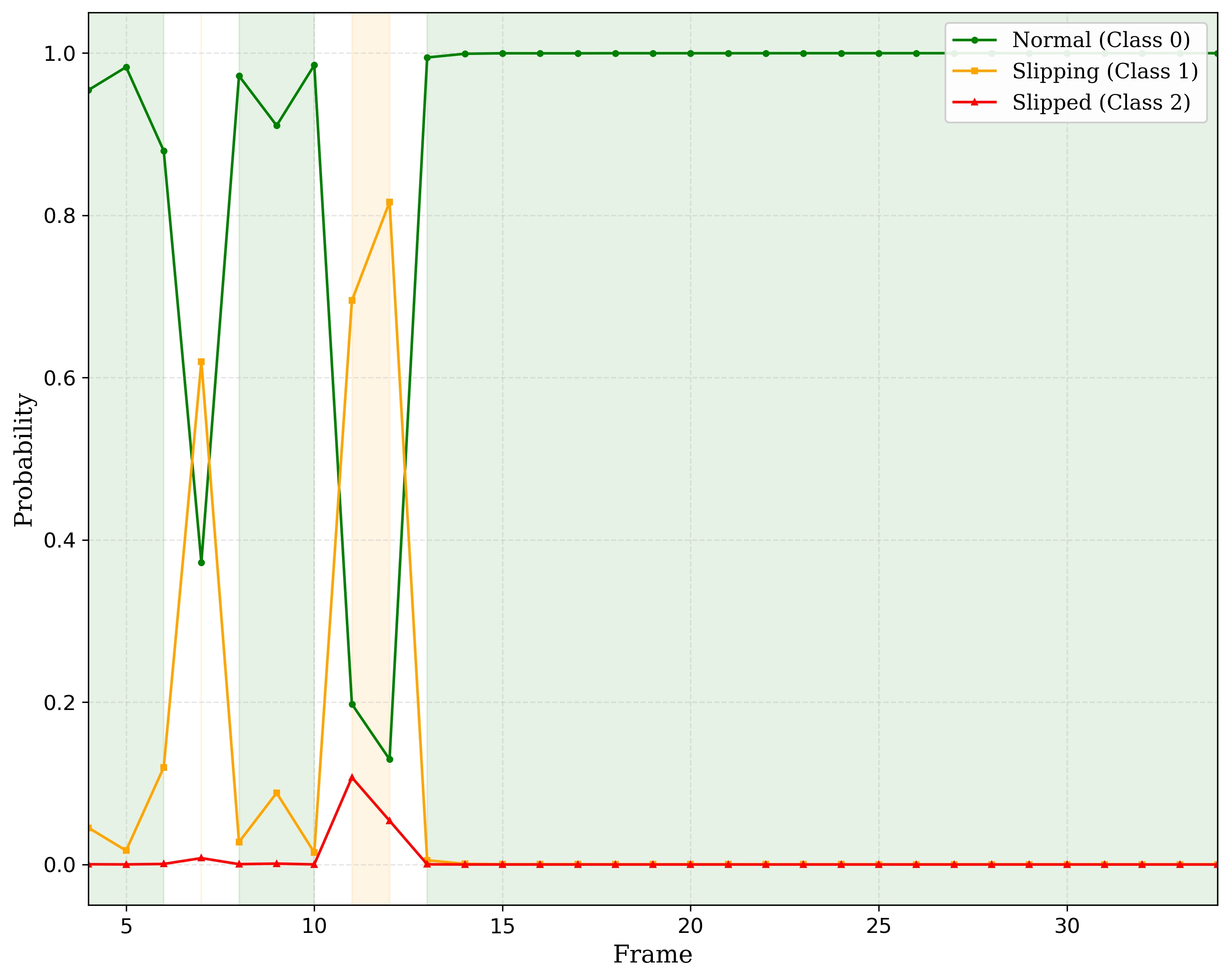}
  \caption{LSTM classifier results. (Left) Video frames 3-14; (Right) Predicted probabilities from the LSTM model. Note that frames were counted starting from 0.}
  \label{fig:lstm_construct}
\end{figure*}

Visualization results of the LSTM classifier on robots in the orchard were presented in Fig. \ref{fig:lstm_construct}. The left panel displayed frames 3-14 with the current strawberry status, the prediction probability of three status, and final strawberry status based on time-stability, while the right panel illustrated the LSTM prediction probabilities. The horizontal axis represented the frame index starting from 0, and the vertical axis indicated the predicted probability of strawberry slippage within the next three frames. If the total number of frames did not reach 5, \textcolor{red}{less than 5} frames were displayed as labels in the top-left corner of the images. The LSTM classifier predicted a slipping status at frames 7, 11, and 12, with corresponding probabilities of 0.62, 0.70, 0.82. Based on the time-stability criterion, secondary inflating and snap-off were triggered at frame 12. The probability curve of strawberry slip prediction was plotted for intuitive observation, as shown in the right panel of Fig. \ref{fig:lstm_construct}.  

\subsection{Operational Efficiency Analysis of the Early Abort Strategy}
% To evaluate the effectiveness of the early abort strategy, the original picking time was quantized and analyzed in Table \ref{time}. 
\textcolor{red}{To assess the effectiveness of the early abort strategy, the time spent continuously harvesting about 30 strawberries within the workspace, as well as the duration of each harvesting stage, were recorded as the baseline harvesting times, as detailed in the second column of Table \ref{time}. In the original process, a complete picking cycle required approximately 10.81 $\pm$ 0.39 s, with the inflating and approaching, swallowing, deflating, snap-off, descending, placing and homing stages taking 1.25 $\pm$ 0.10 s, 0.73 $\pm$ 0.00 s, 0.99 $\pm$ 0.00 s, 0.63 $\pm$ 0.00 s, 0.98 $\pm$ 0.28 s, 4.36 $\pm$ 0.10 s, and 1.88 $\pm$ 0.00 s, respectively. }

\begin{table}%[]
\scriptsize
\centering
\caption{Harvesting time of the robot arm (Mean $\pm$ SD).   (unit: s). -- indicated that the execution time at the corresponding stage under a certain fault was identical to the original (baseline) time. $\times$ indicated that the corresponding stage was not executed. }\label{time}
\begin{tabular}{@{}l|c|c|c|c|c}
\toprule
Action& Original &Empty Grasp & Misgrasp & Slipped &Compensation\\
 % Table header row
\midrule
Inflating \& approaching &  1.25 $\pm$ 0.10& -- & --&  -- &--\\
Compensation & $\times$ &  $\times$ & $\times$ & $\times$ & 0.64 $\pm$ 0.24\\

Swallowing & 0.73 $\pm$ 0.00 & -- &  -- & -- & --\\
Deflating& 0.99 $\pm$ 0.00 &0.42 $\pm$ 0.21 &  0.39 $\pm$ 0.16& -- &--\\
Snap-off & 0.63 $\pm$ 0.00  &$\times$&  $\times$& 0.87 $\pm$ 0.15&  1.26 $\pm$ 0.25 \\
Descending & 0.98 $\pm$ 0.28 &--&  --&    --&--\\
Placing & 4.36 $\pm$ 0.10& $\times$&  $\times$& $\times$ & --\\
Homing & 1.88 $\pm$ 0.00  & -- &  --&  --&--\\
\midrule
Total Time & 10.81 $\pm$ 0.39 & 5.01 $\pm$ 0.84&  4.56 $\pm$ 0.19 &  6.47 $\pm$ 0.23&   11.88 $\pm$  0.34\\
\bottomrule
\end{tabular}
\end{table}

\begin{figure}
\scriptsize
    \centering
    \includegraphics[width=0.95\linewidth]{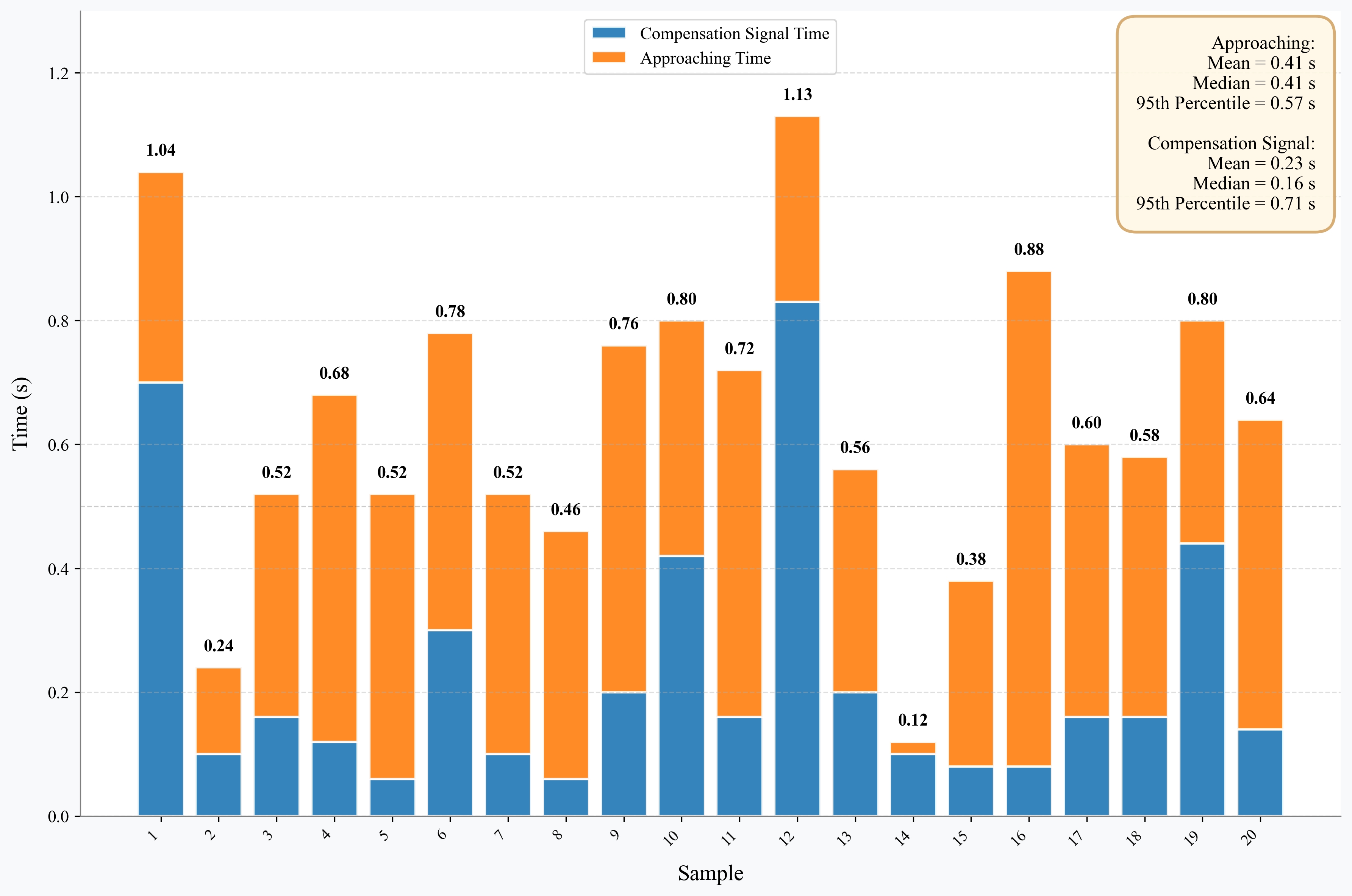}
    \caption{Relative Error Compensation Time}
    \label{fig:compensation-time}
\end{figure}

When the gripper \textcolor{red}{reached the position} beneath the strawberry picking point, the relative error was estimated, and if it exceeded the \textcolor{red}{maximum} threshold, a compensation action was triggered to align the gripper with the strawberry, adding \textcolor{blue}{0.64 $\pm$ 0.24 s} to the cycle. The distribution of compensation signal time and secondary inflating and approaching time were recorded and shown in Fig. \ref{fig:compensation-time}. The minimum and maximum time \textcolor{red}{relative error compensation} were 0.12 s and 1.13 s. 
\textcolor{red}{The total relative error compensation time comprised two distinct phases: the compensation signal reception phase and the gripper's secondary inflation and approach phase. Data from the signal reception phase exhibited a pronounced right-skewed distribution (Median = 0.16 s, Mean = \textcolor{blue}{0.23 s}), indicating its high susceptibility to occasional latency spikes. This vulnerability created a long tail, extending the 95th percentile to 0.71 s. Mechanistically, calculating the relative error fundamentally relied on the simultaneous recognition and continuous tracking of both the end-effector and the target strawberry. However, this visual perception pipeline was sensitive to dynamic illumination changes, which could induce spatial bounding box jitter or momentary tracking failures. Although filtering was applied by constraining the position of the end-effector within the robotic arm's coordinate frame, this perceptual uncertainty directly manifested as physical latency in generating the compensation signal. }

\textcolor{red}{For the gripper's secondary inflation and approach phase, the execution time exhibited a relatively symmetric distribution (Median = 0.41 s, Mean = 0.41 s). This alignment between the mean and median suggested a stable performance of the underlying pneumatic system and kinematic control, indicating minimal software-level latency. Notably, the 95th percentile for this phase was recorded at 0.57 s. This relatively narrow temporal span not only reflected a rapid hardware response but also implied the effectiveness of the system's initial visual localization. It indicated that most relative error compensations required only minor physical displacements, thereby helping to keep the compensation action duration under 0.6 s. }

For grasp adjustment, the response times for empty grasp and misgrasp during the deflating stage were 0.42 $\pm$ 0.21 and 0.39 $\pm$ 0.16, respectively. Compared with original deflating stage, grasp adjustment can save approximately 0.5 s. The snap-off and placing stages were skipped, which saved time in these two phases. 
\textcolor{red}{The total time spent on empty grasp and misgrasp was $5.01 \pm 0.84$ s and $4.56 \pm 0.19$ s, respectively. The discrepancy in total time between these two scenarios, reflected in both the mean and the standard deviation, was primarily attributed to the distance from the picking point to the robotic arm's home position, which primarily affected the inflating and approaching stage and the descending stage. Compared with baseline method, the mitigation of grasp adjustment reduced the harvesting cycle time by approximately 5 s, thereby improving overall harvesting efficiency in Table \ref{time}. }

\begin{figure}
\scriptsize
    \centering
    \includegraphics[width=0.48\linewidth]{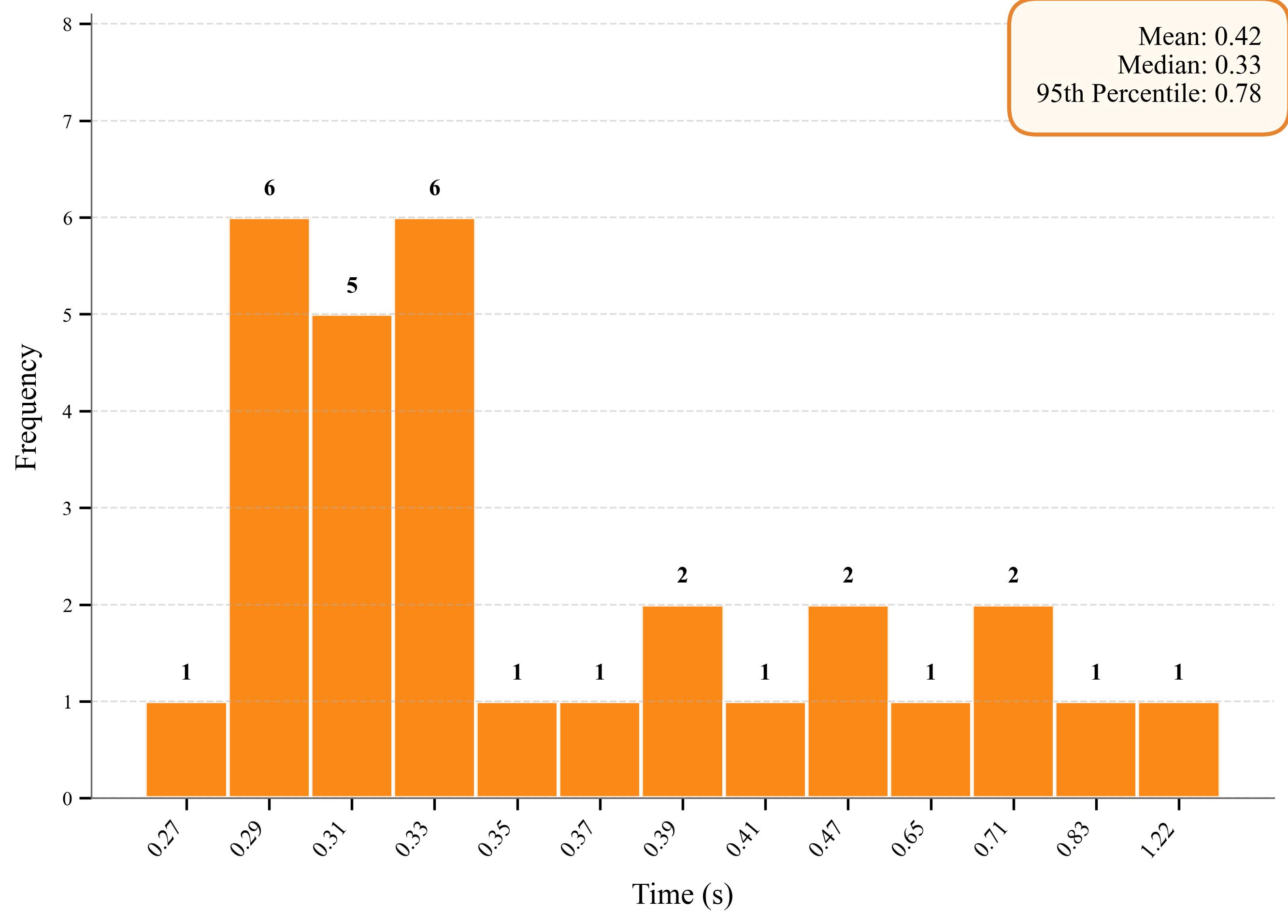}
    \includegraphics[width=0.48\linewidth]{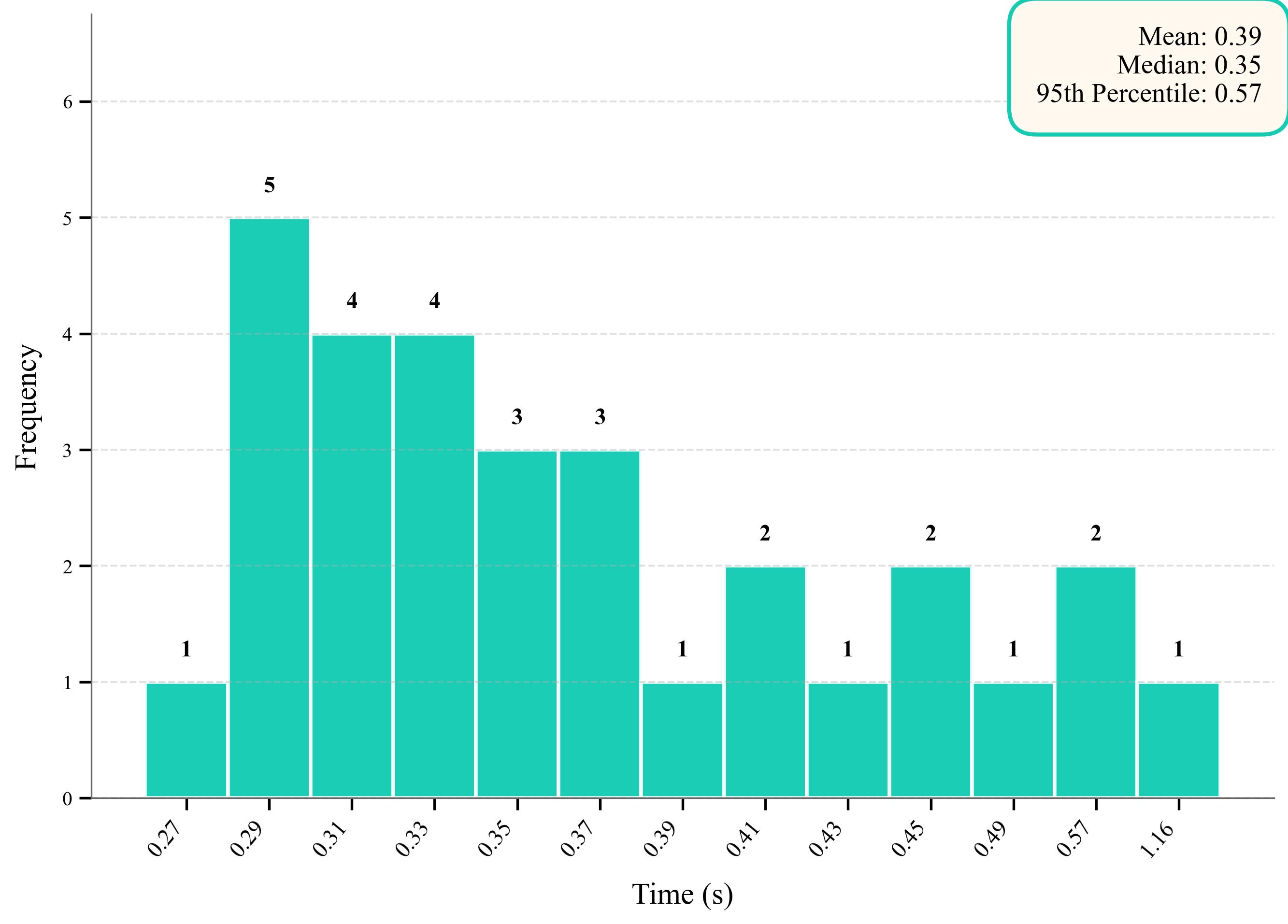}
    \caption{Response time of grasp adjustment during the deflating stage (unit: s). \textcolor{red}{(Left) empty grasp; (right) misgrasp;}}
    \label{fig:earlyabort}
\end{figure}

\textcolor{red}{The response time distributions for empty grasp times and misgrasp were shown in Fig. \ref{fig:earlyabort}.
For the empty grasp, the execution time exhibited a right-skewed distribution. The median response time was highly efficient at $0.33$ s, representing the grasp adjustment baseline performance under typical conditions. However, critical boundary conditions—such as when only the tip of the strawberry was positioned inside the end-effector while the stem end remained outside—introduced extreme values (a long tail), which inflated the mean time to $0.42$ s. Despite these variations, the 95th percentile was maintained at $0.78$ s, demonstrating that the empty grasp guaranteed completion within a sub-second timeframe for the vast majority (95\%) of operations, thereby ensuring overall task fluency. 
Similarly, for the misgrasp, the system demonstrated a highly efficient quick-abort mechanism. The median execution time was recorded at $0.35$ s, indicating rapid abort of unsuccessful attempts. Although a slight right-skew was observed (resulting in a mean of $0.39$ s), the 95th percentile was tightly bounded at $0.57$ s. This remarkably low 95th percentile proved that the system effectively prevents prolonged failure states, aborting 95\% of erroneous grasps within half a second to quickly prepare for the next harvesting cycle. Overall, the majority of the response times for empty grasp and misgrasp were less than 0.5 s.
} 

\textcolor{red}{In Table \ref{tab:slipped-time}, without strawberry slippage prediction, the time spent on the snap-off stage was $0.63 \pm 0.00$ s.
For strawberry slipping that triggered re-inflating and re-snap off, the execution time for successfully harvested strawberries was 1.26 $\pm$ 0.25 s. 
During a secondary snap-off attempt, the re-inflation and snap-off motions were identical to the original ones. Compared to the original snap-off stage, predicting slipping state increased the average execution time by 0.63 s during the snap-off. In other words, this 0.63 s increment represented the combined duration of waiting for the slipping signal and executing the gripper's retraction. For slipped, the total execution time was $0.87 \pm 0.15$ s during snap-stage. This duration represented the scenario where a strawberry slipping and ultimately slipped again despite a secondary re-inflation and snap-off attempt. When the early abort strategy was triggered due to strawberry slipped, the total execution time of the harvesting cycle was reduced to 6.47 $\pm$ 0.23 s. Compared with the original cycle time of 10.81 $\pm$ 0.39 s, the slipped abort strategy reduced the harvesting cycle time by approximately 4 s. The picking cycle time increased to 11.88 $\pm$ 0.34 s after integrating relative error compensation and a secondary snap-off, compared with the original cycle time, representing an increase of approximately 1 s.}

\begin{table}
\scriptsize
    \centering
    \caption{Robot arm response time of strawberry slip prediction (unit: s). \textcolor{red}{Slipping time denoted the duration of the snap-off stage when the secondary re-inflation and snap-off attempt successfully harvested the strawberry; slipped  time referred to the duration when the strawberry ultimately fell and failed to be harvested despite the secondary attempt. }}
    \begin{tabular}{l|ccc}
    \toprule
        Strawberry Status & Normal & Slipping & Slipped\\
        \midrule
        Time (s) &  0.63 $\pm$ 0.00 & 1.26 $\pm$ 0.25&  0.87 $\pm$ 0.15\\
        \bottomrule
    \end{tabular}
    \label{tab:slipped-time}
\end{table}

Finally, the success rates of grasp adjustment and strawberry slippage prediction were assessed through repeated trials based on the robotic arm’s response. As shown in Table \ref{tab:success-rate}, the system achieved an approximately 90.00\% success rate in identifying empty grasp or grasp of unripe strawberry, thereby ensuring appropriate corrective actions on the robot arm. For strawberry slip prediction during the snap-off stage, the success rate for strawberries in a normal state reached 96.88\%. For slipping, the success rate of \textcolor{red}{secondary snap-off attempt} was 81.25\%, while slipped was 88.89\%. \textcolor{red}{Notably, the term slipped referred to the state in which the strawberry tip was still held within the end-effector, but the main body of the fruit was nearly completely outside. It did not encompass situations where the strawberry was entirely out of the micro-camera's field of view. }The relatively lower success rate for initiating secondary inflation and snap-off during snap-off was mainly attributed to the short duration of the slip event, which requires rapid prediction within a limited time window. In addition, performance was influenced by the initial grasping posture of the end-effector during the first contact with the strawberry.

\begin{table}
\scriptsize
    \centering
    \caption{Statistics on robotic arm response success rate for repeated grasp judgments and slip prediction. \textcolor{red}{Slipped denoted the condition where the tip of the strawberry remained inside the end-effector, yet the bulk of the fruit was almost entirely outside the gripper. Notably, this definition excluded cases where the strawberry became completely invisible to the micro-camera.}}
    \begin{tabular}{l|cc|ccc}
    \toprule
     \multirow{2}{*}{Result}& \multicolumn{2}{c|}{Grasp Adjustment} & \multicolumn{3}{c}{Strawberry slip prediction} \\
          & Empty Grasp & Misgrasp & Normal & Slipping & Slipped \\
      \midrule
         Success & 27 & 30 & 31 & 26 & 32\\
        Failure & 3 & 3 & 1 & 6 & 4 \\
        \midrule
        Success Rate & 90.00\% & 90.91\% & 96.88\% & 81.25\% & 88.89\% \\
        \bottomrule
    \end{tabular}
    
    \label{tab:success-rate}
\end{table}

\section{Discussion}
Vision-based fault diagnosis and self-recovery offered an effective means of enhancing the stability and efficiency of strawberry-harvesting robots. During the inflating and approaching stage, the positional relationship between the gripper and the harvesting point served as a visual indicator of accumulated errors caused by fruit recognition, localization, hand–eye calibration, inverse kinematics, \textcolor{red}{and fruit motion.} Without requiring an additional camera, relative errors based on the simultaneous target-gripper detection were computed to align the strawberry with the end-effector. For early abort feedback during the deflating and snap-off stages, a micro-optical camera was embedded in the end-effector to detect and predict the probability of strawberry slippage in the gripper. This approach compensated for the absence of force feedback in the flexible pneumatic gripper, whose deformation during inflation and deflation rendered conventional force sensing impractical. Mechanical failures, inverse kinematics errors, and other control-related faults were not addressed in this work.

Additionally, self-learning, continual learning and adaptive evolutionary perception, planning and decision-making can be applied to reduce dataset dependency and improve the generalization and robustness of the strawberry-harvesting robot. With the rapid development of embodied artificial intelligence and robotic agents, a new wave of end-to-end, multi-modal, large-model-based perception, self-planning and self-decision frameworks is emerging in the robotics domain. These advances offer valuable insights for building strawberry harvesting robots and enabling collaborative operations among multiple robots. In the future, developing end-to-end active learning and continuously evolving methods is expected to become a major research trend.

\section{Conclusion}
This paper proposed a visual-based fault diagnosis and self-recovery system to address gripper offset and strawberry slippage during harvesting. \textcolor{red}{An efficient end-to-end multi-task perception network, SRR-Net, was developed, featuring a simultaneous target-gripper detection mechanism to compensate for relative error.} 
To monitor the strawberry’s status in the gripper, a miniature optical camera was embedded at its base. \textcolor{red}{A MobileNet V3-Small classifier was adapted to evaluate the grasp status during the deflating stage.} For strawberry slippage during the snap-off stage, \textcolor{red}{a time-series LSTM classifier was applied to predict the slippage status of the strawberry within the gripper.} Experimental results demonstrated that the mean absolute errors of compensation mechanism reduced \textcolor{red}{mean absolute} errors to 3.12 mm and \textcolor{blue}{4.06 mm}. \textcolor{red}{The grasp adjustment saved approximately 0.5 s during the deflating stage and reduced the total harvesting cycle time by about 5 s. The strawberry slip prediction improved the harvesting success rate for slipping strawberries, enabling 81.25\% of them to be successfully harvested. Furthermore, for slipped strawberries, the early abort strategy saved approximately 4 s compared to the total harvesting cycle time.}
Overall, the proposed system effectively improved the efficiency of strawberry harvesting. 

\bibliographystyle{elsarticle-harv}  % 或 elsarticle-num
\bibliography{bib2rd}  % 不加 .bib 扩展名
 
\section*{CRediT authorship contribution statement}
Meili Sun: Methodology, Software, Validation, Writing - original draft. Chunjiang Zhao: Investigation, Funding acquisition, Writing - review editing. Lichao Yang: Data curation, Methodology, Software, Visualization. Hao Liu: Data curation, Methodology, Software. Shimin Hu: Data curation, Software, Visualization. Ya Xiong: Conceptualization, Methodology, Investigation, Funding acquisition, Writing - review editing.

\section*{Declaration of competing interest}
\textcolor{blue}{Chunjiang Zhao is the editor-in-chief for Artificial Intelligence in Agriculture and was not involved in the editorial review or the decision to publish this article. All authors declare that there are no competing interests.}

\section*{Declaration of generative AI and AI-assisted technologies in the manuscript preparation process}

During the preparation of this work the authors used ChatGPT in order to improve the language and readability. After using this tool/service, the authors reviewed and edited the content as needed and take full responsibility for the content of the published article.

\section*{Acknowledgments} 
This work was supported by the Beijing Academy of Agricultural Artificial Intelligence and Robotics - Key Technology Research of Strawberry Harvesting Robot Incorporating Visual-Force Perception and Humanoid Cooperative Operation, the Haidian District Bureau of Agriculture and Rural Affairs, the Innovation Ability Project of BAAFS (KJCX20240321), the BAAFS Foundation for Excellent Young Scientists (Grant No. YKPY2025007) and the National Natural Science Foundation of China (NSFC) Excellent Young Scientists Fund (overseas).
\end{document}